\documentclass[10pt,A4paper]{journal}

\usepackage{bm, amssymb, mathtools, makecell, color, eqnarray, float, booktabs,fancyhdr}
\usepackage[square,numbers]{natbib}

\usepackage[left=0.5in, right=.5in, top=1.0in ]{geometry}

\newcommand{\ce}{constrained expression}
\newcommand{\ces}{constrained expressions}
\newcommand{\tfc}{\emph{Theory of  Functional Connections}}
\newcommand{\B}[1]{{\bm #1}}
\newcommand{\ds}{\displaystyle}
\newcommand{\dd}{\; \text{d}}
\newcommand{\T}{^{\mbox{\tiny T}}}
\newcommand{\Ts}{^{\,\mbox{\tiny T}}}
\newcommand{\funding}[1]{
\vspace{6pt}\noindent{\fontsize{9}{9}\selectfont\textbf{Funding:} {#1}\par}}
\newcommand{\abbreviations}[1]{%
\vspace{12pt}\noindent{\selectfont\textbf{Abbreviations}\par\vspace{6pt}\noindent {\fontsize{9}{9}\selectfont #1}\par}}
\def\@appendixtitles{}
\newcommand{\appendixtitles}[1]{\gdef\@appendixtitles{#1}}

\begin{document}
\title{Analytically Embedding Differential Equation Constraints into Least Squares Support Vector Machines using the Theory of  Functional Connections}
\author{Carl Leake\thanks{Department of Aerospace Engineering, Texas A\&M University, College Station, TX 77843, USA: E-mail: \mbox{leakec@tamu.edu}}, Hunter Johnston\thanks{Department of Aerospace Engineering, Texas A\&M University, College Station, TX 77843, USA: E-mail: \mbox{hunterjohnston@tamu.edu}}, Lidia Smith\thanks{Mathematics Department, Blinn College, Bryan, TX 77802, USA: E-mail: \mbox{lidia.smith@gmail.com}}, Daniele Mortari\thanks{Department of Aerospace Engineering, Texas A\&M University, College Station, TX 77843, USA: E-mail: \mbox{mortari@tamu.edu}} }

\maketitle
\pagestyle{fancy}
\fancyhf{}
\cfoot{\thepage}
\renewcommand{\headrulewidth}{0pt}
\thispagestyle{fancy}

\abstract{Differential equations (DEs) are used as numerical models to describe physical phenomena throughout the field of engineering and science, including heat and fluid flow, structural bending, and systems dynamics. While there are many other techniques for finding approximate solutions to these equations, this paper looks to compare the application of the \tfc\ (TFC) with one based on least-squares support vector machines (LS-SVM). The TFC method uses a \ce, an expression that always satisfies the DE constraints, which transforms the process of solving a DE into solving an unconstrained optimization problem that is ultimately solved via least-squares (LS). In addition to individual analysis, the two methods are merged into a new methodology, called constrained SVMs (CSVM), by incorporating the LS-SVM method into the TFC framework to solve unconstrained problems. Numerical tests are conducted on four sample problems: One first order linear ordinary differential equation (ODE), one first order nonlinear ODE, one second order linear ODE, and one two-dimensional linear partial differential equation (PDE). Using the LS-SVM method as a benchmark, a speed comparison is made for all the problems by timing the training period, and an accuracy comparison is made using the maximum error and mean squared error on the training and test sets. In general, TFC is shown to be slightly faster (by an order of magnitude or less) and more accurate (by multiple orders of magnitude) than the LS-SVM and CSVM approaches.}

\section{Introduction}
Differential equations (DE) and their solutions are important topics in science and engineering. The solutions drive the design of predictive systems models and optimization tools. Currently, these equations are solved by a variety of existing approaches with the most popular based on the Runge-Kutta family \cite{RK}. Other methods include those which leverage low-order Taylor expansions, namely Gauss-Jackson \cite{GJ} and Chebyshev-Picard iteration \cite{MCPI1,MCPI2,MCPI3}, which have proven to be highly effective. More recently developed techniques are based on spectral collocation methods \cite{Spec_Collo}. This approach discretizes the domain about collocation points, and the solution of the DE is expressed by a sum of ``basis'' functions with unknown coefficients that are approximated in order to satisfy the DE as closely as possible. Yet, in order to incorporate boundary conditions, one or more equations must be added to enforce the constraints.

The \tfc\ (TFC) is a new technique that analytically derives a constrained expression which satisfies the problem's constraints exactly while maintaining a function that can be freely chosen \cite{TFC}. This theory, initially called ``Theory of Connections'', has been renamed for two reasons. First, the ``Theory of Connections'' already identifies a specific theory in differential geometry, and second, what this theory is actually doing is ``functional interpolation'', as it provides all functions satisfying a set of constraints in terms of a function and any derivative in rectangular domains of $n$-dimensional spaces. This process transforms the DE into an unconstrained optimization problem where the free function is used to search for the solution of the DE. Prior studies \cite{LDE,NDE,constraints,ToCselectApplications}, have defined this free function as a summation  of basis functions; more specifically, orthogonal polynomials. 

This work was motivated by recent results that solve ordinary DEs using a least-squares support vector machine (LS-SVM) approach \cite{LS_SVP}. While this article focuses on the application of LS-SVMs to solve DEs, the study and use of LS-SVMs remains relevant in many areas. In reference \cite{concrete_tiles} the authors use the support vector machines to predict the risk of mold growth on concrete tiles. The mold growth on roofs affects the dynamics of heat and moisture through buildings. The approach leads to reduced computational effort and simulation time. The work presented in reference \cite{water_runoff} uses LS-SVMs to predict annual runoff in the context of water resource management. The modeling process starts with building a stationary set of runoff data based on mode functions which are used as input points in the prediction by the SVM technique when chaotic characteristics are present. Furthermore, reference~\cite{breakwaters_stability} uses the technique of LS-SVMs as a less costly computational alternative that provides superior accuracy compared to other machine learning techniques in the civil engineering problem of predicting the stability of breakwaters. The LS-SVM framework was applied to tool fault
diagnosis for ensuring manufacturing quality \cite{fault_diagnosis}. In this work, a fault diagnosis method was proposed based on stationary subspace analysis (SSA) used to generate input data used for training with LS-SVMs.

In this article, LS-SVMs are incorporated into the TFC framework as the free function, and the combination of these two methods is used to solve DEs. Hence, the contributions of this article are twofold: (1) This article demonstrates how boundary conditions can be analytically embedded, via TFC, into machine learning algorithms and (2) this article compares using a LS-SVM as the free function in TFC with the standard linear combination of CP. Like vanilla TFC, the SVM model for function estimation \cite{Vapnik} also uses a linear combination of functions that depend on the input data points. While in the first uses of SVMs the prediction for an output value was made based on a linear combination of the inputs $x_i$, a later technique uses a mapping of the inputs to feature space, and the model SVM becomes a linear combination of feature functions $\B{\varphi}(x)$.  Further, with the kernel trick, the function to be evaluated is determined based on a linear combination of kernel functions; Gaussian kernels are a popular choice, and are used in this article. 

This article compares the combined method, referred to hereafter as CSVM for constrained LS-SVMs, to vanilla versions of TFC \cite{LDE,NDE} and LS-SVM \cite{LS_SVP} over a variety of DEs. In all cases, the vanilla version of TFC outperforms both the LS-SVM and the CSVM methods in terms of accuracy and speed. The CSVM method does not provide much accuracy or speed benefit over LS-SVM, except in the PDE problem, and in some cases has a less accurate or slower solution. However, in every case the CSVM satisfies the boundary conditions of the problem exactly, whereas the vanilla LS-SVM method solves the boundary condition with the same accuracy as the remainder of the data points in the problem. Thus, this article provides support that in the application of solving DEs, CP are a better choice for the TFC free function than LS-SVMs. 

While the CSVM method underperforms vanilla TFC when solving DEs, its implementation and numerical verification in this article still provides an important contribution to the scientific community. CSVM demonstrates that the TFC framework provides a robust way to analytically embed constraints into machine learning algorithms; an important problem in machine learning. This technique can be extended to any machine learning algorithm, for example deep neural networks. Previous techniques have enforced constraints in deep neural networks by creating parallel structures, such as radial basis networks \cite{NnRbfConstraints}, adding the constraints to the loss function to be minimized \cite{ConstrainedCNN}, or by modifying the optimization process to include the constraints \cite{ConstrainedNnOptimization}. However, all of these techniques significantly modify the deep neural network architecture or the training process. In contrast, embedding the constraints with TFC does not require this. Instead, TFC provides a way to analytically embed these constraints into the deep neural network. In fact, any machine learning algorithm that is differentiable up to the order of the DE can be seamlessly incorporated into TFC. Future work will leverage this benefit to analyze the ability to solve DEs using other machine learning~algorithms.

\section{Background on the \tfc}
The \tfc\ (TFC) is a generalized interpolation method, which provides a mathematical framework to analytically embed constraints. The univariate approach \cite{TFC} to derive the expression for all functions satisfying $k$ linear constraints follows,
\begin{equation}\label{Eq:gen_con}
    f (t) = g (t) + \ds\sum^k_{i = 1} \eta_i \, p_i (t),
\end{equation}
where $g(t)$ represents a ``freely chosen'' function, $\eta_i$ are the coefficients derived from the $k$ linear constraints, and $p_i(t)$ are user selected functions that must be linearly independent from $g(t)$. Recent research has applied this technique to embedding DE constraints using Equation (\ref{Eq:gen_con}), allowing for least-squares (LS) solutions of initial-value (IVP), boundary-value (BVP), and multi-value (MVP) problems on both linear \cite{LDE} and nonlinear \cite{NDE} ordinary differential equations (ODEs). In general, this approach has developed a fast, accurate, and robust unified framework to solve DEs. The application of this theory can be explored for a second-order DE such that,
\begin{equation}\label{eq:de}
    F(t,y,\dot{y},\ddot{y}) = 0 \qquad \text{subject to:} \quad \begin{cases} y (t_0) = y_0 \\ \dot{y} (t_0) = \dot{y}_0 \end{cases}
\end{equation}

By using Equation (\ref{Eq:gen_con}) and selecting $p_1(t) = 1$ and $p_2 (t) = t$, the \ce\ becomes,
\begin{equation}\label{eq:ivp_ce}
    y(t) = g(t) + \eta_1 + \eta_2 \, t.
\end{equation}

By evaluating this function at the two constraint conditions a system of equations is formed in terms of $\eta$,
\begin{equation*}
    \begin{Bmatrix} y_0 \\ \dot{y}_0 \end{Bmatrix} = \begin{bmatrix} 1 & t_0 \\ 0 & 1\end{bmatrix}
    \begin{Bmatrix} \eta_1 \\ \eta_2 \end{Bmatrix}
\end{equation*}
which can be solved for by matrix inversion leading to,
\begin{equation*}
    \begin{aligned}
    \eta_1 &= (y_0 - g_0) - t_0 \, (\dot{y}_0 - \dot{g}_0) \\
    \eta_2 &= \dot{y}_0 - \dot{g}_0.    
    \end{aligned}
\end{equation*}

These terms can are substituted in Equation (\ref{eq:ivp_ce}) and the final \ce\, is realized,
\begin{equation*}
    y (t) = g (t) + (y_0 - g_0) + (t - t_0) (\dot{y}_0 - \dot{g}_0),
\end{equation*}

By observation, it can be seen that the function for $y(t)$ always satisfies the initial value constraints regardless of the function $g(t)$. Substituting this function into our original DE specified by Equation (\ref{eq:de}) transforms the problem into a new DE with no constraints,
\begin{equation}\label{eq:tilde_de}
    \tilde{F}(t,g,\dot{g},\ddot{g}) = 0.
\end{equation}

Aside from the independent variable $t$, this equation is only a function of the unknown function $g(t)$. By discretizing the domain and expressing $g(t)$ as some universal function approximator, the problem can be posed as an unconstrained optimization problem where the loss function is defined by the residuals of the $\tilde{F}$ function. Initial applications of the TFC method to solve DEs \cite{LDE,NDE} expanded $g(t)$ as some basis (namely Chebyshev or Legendre orthogonal polynomials); however, the incorporation of a machine learning framework into this free function has yet to be explored. This will be discussed in following sections. The original formulation expressed $g(t)$ as,
\begin{equation*}
    g (t) = \B{\xi} \Ts \B{h}(x) \qquad \text{where} \qquad x = x(t),
\end{equation*}
where $\B{\xi}$ is an unknown vector of $m$ coefficients and $\B{h}(x)$ is the vector of $m$ basis functions. In general the independent variable is $t \in [t_0, t_f]$ while the orthogonal polynomials are defined in $x \in [-1, +1]$. This gives the linear mapping between $x$ and $t$,
\begin{equation*}
x = x_0 + \frac{x_f - x_0}{t_f - t_0}(t - t_0) \qquad \leftrightarrow \qquad t = t_0 + \frac{t_f - t_0}{x_f - x_0}(x - x_0).
\end{equation*}

Using this mapping, the derivative of the free function becomes,
\begin{equation*}
\frac{\dd g}{\dd t} =  \frac{\dd \B{\xi}\Ts \B{h}(x)}{\dd t} = \B{\xi}\Ts \frac{\dd \B{h}(x)}{\dd x} \cdot \frac{\dd x}{\dd t},
\end{equation*}
where it can be seen that the term $\frac{\dd x}{\dd t}$ is a constant such that,
\begin{equation*}
    c:= \frac{x_f - x_0}{t_f - t_0}.
\end{equation*}

Using this definition, it follows that all subsequent derivatives are,
\begin{equation*}
    \frac{\dd ^k g}{\dd t^k} = c^k \B{\xi}\Ts \frac{\dd ^k\B{h}(x)}{\dd x^k}.
\end{equation*}

Lastly, the DE given by Equation (\ref{eq:tilde_de}) is discretized over a set of $N$ values of $t$ (and inherently $x$). When using orthogonal polynomials, the optimal point distribution (in terms of numerical efficiency) is provided by collocation points \cite{colloc1,colloc2}, defined as,
\begin{equation*}
	x_i  = -\cos\left(\dfrac{i \pi}{N}\right)  \qquad \text{for} \qquad i = 0,1,\cdots,N,
\end{equation*}

By discretizing the domain, Equation (\ref{eq:tilde_de}) becomes a function solely of the unknown parameter $\B{\xi}$,
\begin{equation}\label{eq:final_de}
    \tilde{F}(\B{\xi}) = 0,
\end{equation}
which can be solved using a variety of optimization schemes. If the original DE is linear then the new DE defined by Equation (\ref{eq:final_de}) is also linear. In this case, Equation (\ref{eq:final_de}) is a linear system,
\begin{equation*}
    A \, \B{\xi} = \B{b},
\end{equation*}
which can be solved using LS \cite{LDE}. If the DE is nonlinear, a nonlinear LS approach is needed, which requires an initial guess for $\B{\xi}_0$. This initial guess can be obtained by a LS fitting of a lower order integrator solution, such as one provided by a simple improved Euler method. By defining the residuals of the DE as the loss function $\mathcal{L}_k := \tilde{F}(\B{\xi}_k)$, the nonlinear Newton iteration is,
\begin{equation*}
     \B{\xi}_{k+1}= \B{\xi}_k - (\mathcal{J}_k\Ts \mathcal{J}_k)^{-1} \mathcal{J}_k\Ts \mathcal{L}_k \quad \text{where} \quad \mathcal{J}_k := \left[\frac{\partial \mathcal{L}}{\partial \B{\xi}}\right]_k
\end{equation*}
where $k$ is the iteration. The convergence is obtained when the $L_2$-norm of $\mathcal{L}$  satisfies $L_2 [\mathcal{L}_k] < \varepsilon$, where $\varepsilon$ is a specified convergence tolerance. The final value of $\B{\xi}$ is then used in the \ce\ to provide an approximated analytical solution that perfectly satisfies the constraints. Since the function is analytical, the solution can be then used for further manipulation (e.g., differentiation, integration, etc.). The process to solve PDEs follows a similar process with the major difference involving the derivative of the \ce. The TFC extension to $n$-dimensions and a detailed explanation of the derivation of these \ces\ are provided in references \cite{M-ToC,M-TFC-Conference}. Additionally, the free function also becomes multivariate, increasing the complexity when using CPs.

\section{ The Support Vector Machine Technique}
\subsection{An Overview of SVMs}
Support vector machines (SVMs) were originally introduced to solve classification problems~\cite{Vapnik}. A classification problem consists of determining if a given input, $x$, belongs to one of two possible classes. The proposed solution was to find a decision boundary surface that separates the two classes. The equation of the separating boundary depended only on a few input vectors called the support~vectors.

The training data is assumed to be separable by a linear decision boundary. Hence, a separating hyperplane, $H$, with equation $\B{w}\T \B{\varphi} (x) + b = 0$, is sought. The parameters are rescaled such that the closest training point to the hyperplane $H$, let's say  $(x_k, y_k)$, is on a parallel hyperplane $H_1$ with equation $\B{w}\T \B{\varphi} (x) + b = 1$. By using the formula for orthogonal projection, if $\B{x}$ satisfies the equation of one of the hyperplanes, then the signed distance from the origin of the space to the corresponding hyperplane is given by $\B{w}\T\B{\varphi} (x) /\B{w}\T\B{w}$. Since $\B{w}\T\B{\varphi} (x) $ equals $-b$ for $H$, and $1-b$ for $H_1$, it follows that the distance between the two hyperplanes, called the ``separating margin'', is $1/\B{w}\T\B{w}$. Thus to find the largest separating margin, one needs to minimize $ \B{w}\T\B{w}$. The optimization problem becomes,
\begin{equation*}
    \min \dfrac{1}{2} \left(\B{w}\T \B{w} \right) \quad \text{subject to: } \, y_i(\B{w}\T \B{\varphi} (x_i) + b) \geq 1, \quad i =1,\dots,m.
\end{equation*}

If a separable hyperplane does not exist, the problem is reformulated by taking into account the classification errors, or slack variables, $\xi_i$ , and a linear or quadratic expression is added to the cost function.  The optimization problem in the non-separable case is, 
\begin{equation*}
    \min \dfrac{1}{2} \left(\B{w}\T \B{w} \right) +C \left(\sum \xi_i\right) \quad \text{subject to: } \, y_i(\B{w}\T \B{\varphi} (x_i) + b) \geq 1-\xi_i.
\end{equation*}

When solving the optimization problem by using Lagrange multipliers, the function $\B{\varphi} (t)$ always shows up as a dot product with itself; thus, the kernel trick \cite{KernelTrick} can be applied. In this research, the kernel function chosen is the radial basis function (RBF) kernel proposed in \cite{LS_SVP}. Hence, the function $\B{\varphi} (t)$ can be written using the kernel \cite{KernelTrick},
\begin{equation}\label{eq:Kernel}
    K(t_i,t)=\B{\varphi}(t_i)\T\B{\varphi}(t)=\exp\left(-\dfrac{\left(t-t_i\right)^2}{\sigma^2}\right),
\end{equation}
and its partial derivatives \cite{LS_SVP,Reviewer1Article},
\begin{align}\label{eq:KernelDerivatives}
    K (t_i, t_j) =  \B{\varphi} (t_i)\T \B{\varphi} (t_j) &= \exp\left(-\dfrac{(t_i - t_j)^2}{\sigma^2}\right) \nonumber\\
    K_1 (t_i,  t_j) = \B{\varphi}' (t_i)\T \B{\varphi} (t_j) &= -\dfrac{2(t_i - t_j)}{\sigma^2} \exp\left(-\dfrac{(t_i - t_j)^2}{\sigma^2}\right) \nonumber\\
    K_1\T (t_i, t_j) = \B{\varphi} (t_i)\T \B{\varphi'} (t_j) &= \dfrac{2(t_i - t_j)}{\sigma^2} \exp\left(-\dfrac{(t_i - t_j)^2}{\sigma^2}\right) \\
    K_{11} (t_i, t_j) = \B{\varphi}' (t_i)\T \B{\varphi'} (t_j) &= \dfrac{2}{\sigma^2} - \dfrac{4 (t_i - t_j)^2}{\sigma^4} \exp\left(-\dfrac{(t_i - t_j)^2}{\sigma^2}\right)\nonumber,
\end{align}
where the Kernel bandwidth, $\sigma$, is a tuning parameter that must be chosen by the user. 

We follow the method of solving DEs using RBF kernels proposed in \cite{LS_SVP}. As an example, we take a first order linear initial value problem,
\begin{equation*}
    y'- p (t) y = r (t), \quad \text{subject to:} \quad y (t_0) = y_0,
\end{equation*}
to be solved on the interval $[t_0,t_{f}]$. The domain is partitioned into $N$ sub-intervals using grid points $t_0, t_1, \dots, t_N = t_f$, which from a machine learning perspective represents the training points. The~model, 
\begin{equation}\label{yhat}
    \hat{y}(x) = \sum_{i=1}^N w_i \varphi_i (x) + b = \B{w}\T \B{\varphi}(x) + b
\end{equation}
is proposed for the solution $y(t)$. Note that the number of coefficients $w_i$ equals the number of grid points $t_i$, and thus the system of equations used to solve for the coefficient is a square matrix. Let $\B{e}$ be the vector of residuals obtained when using the model solution $\hat{y} (t)$ in the DE, that is, $e_i$ is the amount by which $\hat{y}(t_i)$ fails to satisfy the DE,
\begin{equation*}
    \hat{y}'(t_i) - p(t_i)\hat{y}(t_i) - r(t_i) = e_i.
\end{equation*}

This results in,
\begin{equation*}
    \B{w}\T\B{\varphi}'(t_i) = p(t_i) [\B{w}\T\B{\varphi}(t_i) + b] + r (t_i) + e_i,
\end{equation*}
and for the initial condition, it is desired that,
\begin{equation*}
    \B{w}\T \B{\varphi} (t_0) + b = y_0,
\end{equation*}
is satisfied exactly. In order to have the model close to the exact solution, the sum of the squares of the residuals, $\B{e}\T\B{e}$, is to be minimized. This expression can be viewed as a regularization term added to the objective of maximizing the margin between separating hyperplanes. The problem is formulated as an optimization problem with constraints,
\begin{equation*}
    \min\dfrac{1}{2} \left(\B{w}\T\B{w} + \gamma \, \B{e}\T\B{e}\right) \quad \text{subject to:} \, \begin{cases}
        \B{w}\T\B{\varphi}'(t_i) - p(t_i)\left(\B{w}\T\B{\varphi}(t_i) + b\right) - r(t_i) - e_i = 0 \\
        \B{w}\T\B{\varphi}(t_0) + b - y_0 = 0.
    \end{cases}
\end{equation*}

Using the method of Lagrange multipliers, a loss function, $\cal L$, is defined using the objective function from the optimization problem and appending the constraints with corresponding Lagrange multipliers $\alpha_i$  and $\beta$.
\begin{equation*}
    \mathcal{L} = \dfrac{1}{2} \left(\B{w}\T\B{w} + \gamma \, \B{e}\T\B{e}\right) + \alpha_i\big[\B{w}\T\B{\varphi}'(t_i) - p(t_i)\left(\B{w}\T\B{\varphi}(t_i) + b\right) - r(t_i) - e_i\big] + \beta\big[\B{w}\T\B{\varphi}(t_0) + b - y_0\big]
\end{equation*}

The values where the gradient of ${\cal L}$ is zero give candidates for the minimum.
\begin{align*}
&\dfrac{\partial{\cal L}}{\partial\B{w}}=0\qquad\to\qquad\B{w}=\sum_{i=1}^N\alpha_i\left[\B{\varphi}'(t_i)-p(t_i)\B{\varphi}(t_i)\right]+\beta\B{\varphi}(t_0)\\
&\dfrac{\partial{\cal L}}{\partial e_i}=0\qquad\to\qquad\gamma e_i=-\alpha_i\\
&\dfrac{\partial{\cal L}}{\partial b}=0\qquad\to\qquad0=\sum_{i=1}^N\alpha_iq(t_i)-\beta\\
&\dfrac{\partial{\cal L}}{\partial\alpha_i}=0\qquad\to\qquad0=\B{w}\T\B{\varphi}'(t_i)-p(t_i)\left(\B{w}\T\B{\varphi}(t_i)+b\right)-g(t_i)-e_i\\
&\dfrac{\partial{\cal L}}{\partial\beta} = 0 \qquad\to\qquad 0 = \B{w}\T\B{\varphi}(t_0) + b - y_0
\end{align*}

Note that the conditions found by differentiating ${\cal L}$ with respect to $\alpha_i$ and $\beta$ are simply the constraint conditions, while the remaining conditions are the standard Lagrange multiplier conditions that the gradient of the function to be minimized is a linear combination of the gradients of the constraints.~Using,
\begin{equation*}
    \B{w} = \ds\sum_{j=1}^N \alpha_j\left[\B{\varphi}'(t_j) - p(t_j)\B{\varphi}(t_j)\right] + \beta \B{\varphi} (t_0),
\end{equation*}
we obtain a new formulation of the approximate solution 
\begin{equation*}
\begin{aligned}
    \hat{y}(t) &= \ds\sum_{j=1}^N \alpha_j\left[\B{\varphi}'(t_j) - p(t_j)\B{\varphi}(t_j)\right]\T\B{\varphi}(t) + \beta \B{\varphi} (t_0)\T\B{\varphi}(t) + b 
\end{aligned}
\end{equation*}
where the inner products of $\B{\varphi}(t)$ can be re-written using Equations \eqref{eq:Kernel} and \eqref{eq:KernelDerivatives}, and the parameter $\sigma$ in the kernal matrix is a value that is learned during the training period together with the coefficients $\B{w}$. The remaining gradients of $\mathcal{L}$ can be used to form a linear system of equations where $\alpha_i$, $\beta$, and $b$ are the only unknowns. Note, that this system of equations can also be expressed using the kernal matrix and its partial derivatives rather than inner-products of $\B{\varphi}$.

\subsection{Constrained SVM (CSVM) Technique}

In the TFC method \cite{TFC}, the general \ce\ can be written for an initial value constraint as,
\begin{equation*}
    y (t) = g (t) + (y_0 - g_0),
\end{equation*}
where $g (t)$ is a ``freely chosen'' function. In prior studies \cite{LDE,NDE,ToCselectApplications}, this free function was defined by a set of orthogonal basis functions, but this function can also be defined using SVMs,
\begin{equation*}
    g (t) = \ds\sum_{i = 1}^N w_i \varphi_i (t) = \B{w}\T \B{\varphi} (t),
\end{equation*}
where $g_0$ becomes,
\begin{equation*}
    g (t_0) = \sum_{i = 1}^N w_i \varphi_i (t_0) = \B{w}\T \B{\varphi} (t_0).
\end{equation*}

This leads to the equation,
\begin{equation}\label{y_TFC}
    y (t) = \B{w}\T \left[\B{\varphi} (t) - \B{\varphi} (t_0)\right] + y_0,
\end{equation}
where the initial value constraint is always satisfied regardless of the values of $\B{w}$ and $\B{\varphi}(t)$. Through this process, the constraints only remain on the residuals and the problem becomes,
\begin{equation*}
    \min\dfrac{1}{2}\left(\B{w}\T\B{w} + \gamma\B{e}\T\B{e}\right) \quad \text{subject to:} \quad \B{w}\T \B{\varphi}' (t_i) - p(t_i) \left[\B{w}\T \B{\varphi} (t_i) - \B{w}\T \B{\varphi} (t_0) + y_0\right] - r (t_i) - e_i = 0.
\end{equation*}

Again, using the method of Lagrange multipliers, a term is introduced for the constraint on the residuals, leading to the expression,
\begin{equation*}
    {\cal L}(\B{w}, \B{e}, \B{\alpha}) = \dfrac{1}{2} \left(\B{w}\T \B{w} + \gamma \B{e}\T \B{e}\right) - 
     \sum_{i=1}^N \alpha_i \big[\B{w}\T \B{\varphi}' (t_i) -
     p (t_i) \left(\B{w}\T \B{\varphi} (t_i) - \B{w}\T \B{\varphi} (t_0) + y_0\right) - r (t_i) - e_i\big].
\end{equation*}

The values where the gradient of ${\cal L}$ is zero give candidates for the minimum,
\begin{equation*}
\begin{aligned}
    \dfrac{\partial{\cal L}}{\partial\B{w}} = 0 &\qquad\to\qquad \B{w} = \sum_{i=1}^N \alpha_i \left[\B{\varphi}' (t_i) - p(t_i) \left(\B{\varphi} (t_i) - \varphi(t_0)\right)\right] \\ 
    \dfrac{\partial{\cal L}}{\partial e_i} = 0 &\qquad\to\qquad e_i = -\frac{\alpha_i}{\gamma} \\
    \dfrac{\partial{\cal L}}{\partial\alpha_i} = 0 &\qquad\to\qquad 0 = \B{w}\T\B{\varphi}' (t_i) -  p(t_i)\left(\B{w}\T\left(\B{\varphi} (t_i) - \B{\varphi} (t_0)\right) + y_0\right) - r (t_i) - e_i.
\end{aligned}
\end{equation*}

Using,
\begin{equation*}
    \B{w} = \ds \sum_{j=1}^N \alpha_j \left[\B{\varphi}' (t_j) - p(t_j) \left(\B{\varphi} (t_j) - \varphi(t_0)\right)\right],
\end{equation*}
we obtain a new formulation of the approximate solution given by Equation (\ref{y_TFC}),  that can be expressed in terms of the kernel and its derivatives. Combining the three equations for the gradients of $\cal L$, we can obtain a linear system with unknowns $\alpha_j$,
\begin{equation*}
    \sum_{j=1}^N M_{ij} \alpha_j = r (t_i) + p (t_i) y_0.
\end{equation*}

The coefficient matrix is given by,
\begin{equation*}
\begin{aligned}
    M_{ij} = K_{11} (t_i, t_j) &- p (t_j)\left[K_1 (t_i, t_j) - K_1 (t_i, t_0)\right]  - p (t_i) K_y (i, j) + \delta_{ij}/\gamma,
\end{aligned}
\end{equation*}
where  we use the notation,
\begin{equation*}
\begin{aligned}
    K_4 (t_i,t_j) &= K (t_i, t_j) - K (t_j, t_0) - K (t_i, t_0) + 1 \\
    K_y (t_i, t_j) & = K_1 (t_j, t_i) - K_1 (t_j, t_0) - p (t_j) K_4 (t_i, t_j).
\end{aligned}
\end{equation*}

Finally, in terms of the kernel matrix, the approximate solution at the grid points is given by,
\begin{equation*}
    y(t_i) = \ds\sum_{j=1}^N \alpha_j K_y (t_i, t_j) + y_0,
\end{equation*}
and a formula for the approximate solution at an arbitrary point $t$ is given by,
\begin{equation*}
    y(t) = \ds\sum_{j=1}^N \alpha_j K_y (t, t_j) + y_0.
\end{equation*}

\subsection{Nonlinear ODEs}

The method for solving nonlinear, first-order ODEs with LS-SVM comes from reference \cite{LS_SVP}. Nonlinear, first-order ODEs with initial value boundary conditions can be written generally using the form,
\begin{equation*}
    y' (t) = f (t, y), \quad y (t_0) = y_0, \quad t \in [t_0, t_f].
\end{equation*}

The solution form is again the one given in Equation (\ref{yhat}) and the domain is again discretized into $N$ sub-intervals, $t_0, t_1, \dots, t_N$ (training points). Let $e_i$ be the residuals for the solution $\hat{y} (t_i)$,
\begin{equation*}
    e_i = \hat{y}' (t_i) - f (t_i, \hat{y} (t_i)).
\end{equation*}

To minimize the error, the sum of the squares of the residuals is minimized. As in the linear case, the regularization term $\B{w}\T \B{w}$ is added to the expression to be minimized. Now, the problem can be formulated as an optimization problem with constraints,
\begin{equation*}
    \min\dfrac{1}{2} \left(\B{w}\T\B{w} + \gamma \, \B{e}\T\B{e}\right)
    \quad \text{subject to:} \, \begin{cases}
    \B{w}\T \B{\varphi}' (t_i) = f (t_i, y_i) + e_i \\
    \B{w}\T \B{\varphi} (t_0) + b = y_0 \\
    y_i = \B{w}\T \B{\varphi} (t_i) + b.
    \end{cases}
\end{equation*}

The variables $y_i$ are introduced into the optimization problem to keep track of the nonlinear function $f$ at the values corresponding to the grid points. The method of Lagrange multipliers is used for this optimization problem just as in the linear case. This leads to a system of equations that can be solved using a multivariate Newton's method. As with the linear ODE case, the set of equations to be solved and the dual form of the model solution can be written in terms of the kernel matrix and its~derivatives. 

The solution for nonlinear ODEs when using the CSVM technique is found in a similar manner, but the primal form of the solution is based on the constraint function from TFC. Just as the linear ODE case changes to encompass this new primal form, so does the nonlinear case. A complete derivation for nonlinear ODEs using LS-SVM and CSVM is provided in Appendix \ref{app:B}.

\subsection{Linear PDEs}
The steps for solving linear PDEs using LS-SVM are the same as when solving linear ODEs, and are shown in detail in reference \cite{SvmPde}. The first step is to write out the optimization problem to be solved. The second is to solve that optimization problem using the Lagrange multipliers technique. The third is to write the resultant set of equations and dual-form of the solution in terms of the kernel matrix and its derivatives. 

Solving linear PDEs using the CSVM technique follows the same solution steps except the primal form of the solution is derived from a TFC constrained expression. A complete derivation for the PDE shown in problem \#4 of the numerical results section using CSVM is provided in Appendix~\ref{app:C}. The main difficulty in this derivation stems from the numerous amount of times the function $\B{\varphi}$ shows up in the TFC constrained expression. As a result, the set of equations produced by taking gradients of $\mathcal{L}$ contain hundreds of kernel matrices and their derivatives. The only way to make this practical (in terms of the derivation and programming the result) was to write the constrained expression in tensor form. This was reasonable to perform for the simple linear PDE used in this paper, but would become prohibitively complicated for higher dimensional PDEs. Consequently, future work will investigate using other machine learning algorithms, such as neural networks, as the free function in the TFC~framework.

\section{Numerical Results}

This section compares the methodologies described in the previous sections on four problems given in references \cite{LS_SVP} and \cite{SvmPde}. Problem \#1 is a first order linear ODE, problem \#2 is a first order nonlinear ODE, problem \#3 is a second order linear ODE, and problem \#4 is a second order linear PDE. All problems were solved in \verb"MATLAB R2018b" (MathWorks, Natick, MA, USA)
 on a Windows 10 operating system running on an Intel\textsuperscript{\textregistered} Core\textsuperscript{\texttrademark} i7-7700 CPU at 3.60GHz and 16.0 GB of RAM. Since all test problems have analytical solutions, absolute error and mean-squared error (MSE) were used to quantify the error of the methods. MSE is defined as,
\begin{equation}
    \text{MSE} =\frac{1}{n} \sum _{i=1}^{n}(y_i-\hat{y}_i)^{2}
\end{equation}
where $n$ is the number of points, $y_i$ is the true value of the solution, and $\hat{y}_i$ is the estimated value of the solution at the $i$-th point.

The tabulated results from this comparison are included in Appendix \ref{apx:data}. A graphical illustration and summary of those tabulated values is included in the subsections that follow, along with a short description of each problem. These tabulated results also include the tuning parameters for each of the methods. For TFC, the number of basis functions, $m$, was found using a grid search method, where the residual of the differential equation was used to choose the best value of $m$. For LS-SVM and CSVM, the kernel bandwidth, $\sigma$, and the parameter $\gamma$ were found using a grid search method for problems \#1, \#3, and \#4. For problem \#2, the value of $\sigma$ for the LS-SVM and CSVM methods was tuned using fminsearch while the value of $\gamma$ was fixed at $10^{10}$ \cite{LS_SVP}. This method was used in problem \#2 rather than grid search because it did a much better job choosing tuning parameters that reduced the error of the solution. For all problems, a validation set was used to choose the best value for $\sigma$ and $\gamma$~\cite{LS_SVP,Reviewer1Article}. It should be noted that the tuning parameter choice affects the accuracy of the solution. Thus, it may be possible to achieve more accurate results if a different method is used to find the value of the tuning parameters. For example, an algorithm that is better suited to finding global optimums, such as a genetic algorithm, may find better tuning parameter values than the methods used here.

\subsection{Problem \#1} 

Problem \#1 is the linear ODE, 
\begin{equation}\label{Eq:Prob1}
\begin{aligned}
    \dot{y} + \left(t + \frac{1 + 3 t^2}{1 + t + t^3}\right) y = t^3 + 2 t + t^2 \frac{1 + 3 t^2}{1 + t + t^3}, \quad \text{subject to: } y (0) = 1, \quad t\in[0, 1]
\end{aligned}
\end{equation}
which has the analytic solution,
\begin{equation*}
    y(t) = \frac{e^{-t^2/2}}{1+ t + t^3} + t^2
\end{equation*}

\begin{figure}[H]
	\centering
    \includegraphics[width=.5\linewidth]{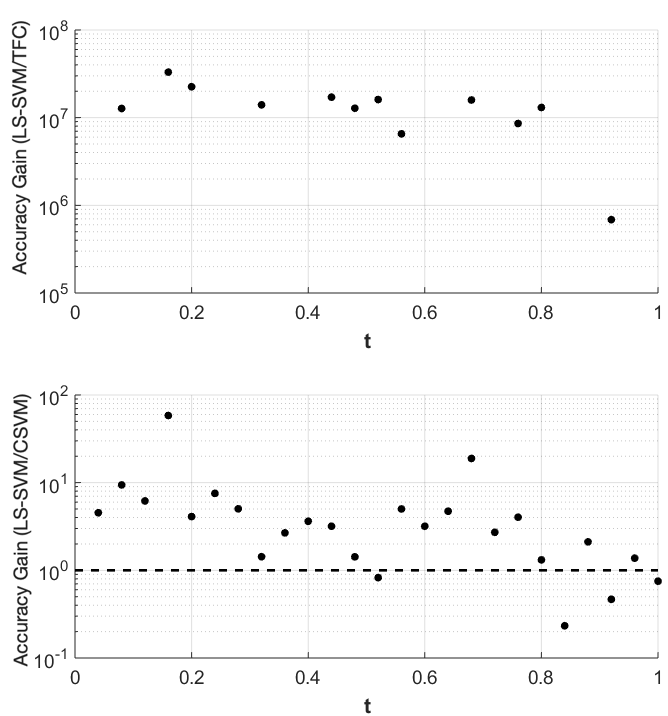}
    \caption{Accuracy gain for the Theory of Functional Connections (TFC) and constrained support vector machine (CSVM) methods over least-squares support vector machines (LS-SVMs) for problem \#1 using 100 training~points.}
    \label{fig:Prob1}
\end{figure}

The accuracy gain of TFC and CSVM compared to LS-SVM for problem \#1 is shown in Figure~\ref{fig:Prob1}. The results were obtained using 100 training points. The top plot shows the error of the LS-SVM solution divided by the error in the TFC solution, and the bottom plot shows the error of the LS-SVM solution divided by the error of the CSVM solution. Values greater than one indicate that the compared method is more accurate than the LS-SVM method, and vice-versa for values less than one.

Figure \ref{fig:Prob1} shows that TFC is the most accurate of the three methods followed by CSVM and finally LS-SVM. The error reduction when using CSVM instead of LS-SVM is typically an order of magnitude or less. However, the error reduction when using TFC instead of the other two methods is multiple orders of magnitude. The attentive reader will notice that the plot that includes TFC solution has less data points in Figure \ref{fig:Prob1} than the other methods. This is because the calculated points and the true solutions vary less than machine level accuracy and when the subtraction operation is used the resulting number becomes~zero.

\begin{figure}[H]
	\centering
    \includegraphics[width=.5\linewidth]{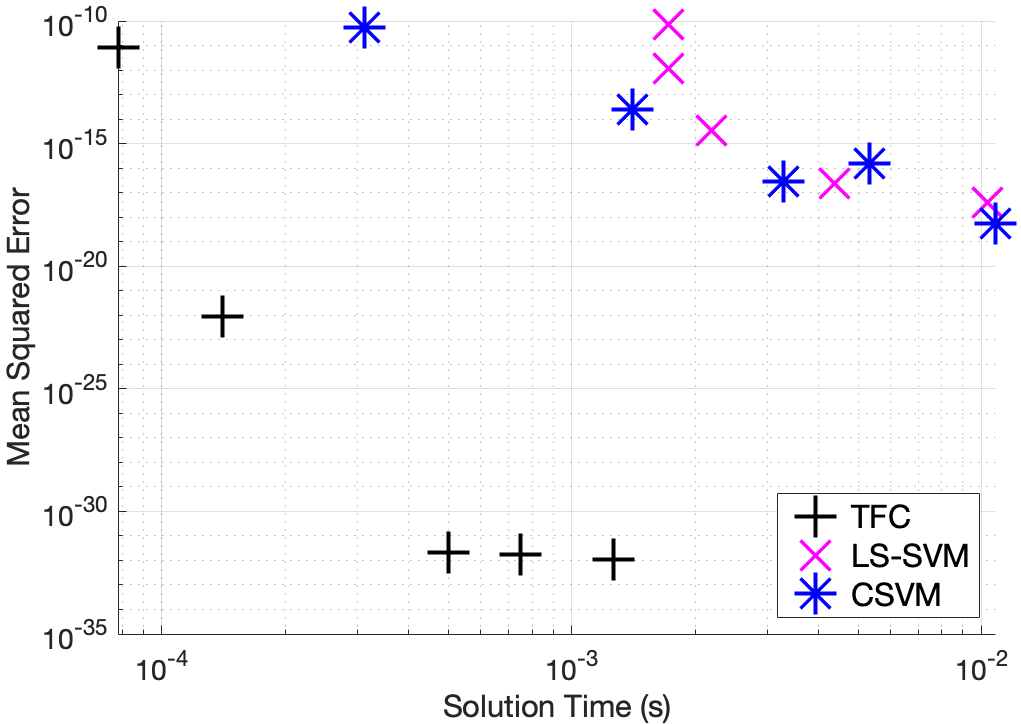}
    \caption{Mean squared error vs. solution time for problem \#1.}
    \label{fig:Prob1TimeAcc}
\end{figure}

Tables \ref{tab:TFC1} through \ref{tab:SVMTFC1} in the appendix compare the three methods for various numbers of training points when solving problem \#1. Additionally, these tables show that TFC provides the shortest training time and the lowest maximum error and mean square error (MSE) on both the training set and test set. The CSVM results are the slowest, but they are more accurate than the LS-SVM results. The accuracy gained when using CSVM compared to LS-SVM is typically less than an order of magnitude. On the other hand, the accuracy gained when using TFC is multiple orders of magnitude. Moreover, the speed gained when using LS-SVM compared to CSVM is typically less than an order of magnitude, whereas the speed gained when using TFC is approximately one order of magnitude. An accuracy versus speed comparison is shown graphically in Figure \ref{fig:Prob1TimeAcc}, where the MSE on the test set is plotted against training time for five specific cases: 8, 16, 32, 50, and 100 training points.

\subsection{Problem \#2} 
Problem \#2 is the nonlinear ODE given by,
\begin{equation}\label{Eq:Prob3}
    \dot{y} = y^2 + t^2, \quad \text{subject to: } y (0) = 1, \quad t \in [0,0.5],
\end{equation}
which has the analytic solution,
\begin{equation*}
    y(t) =  -\frac{t \left(\Gamma \left(\frac{1}{4}\right) J_{-\frac{3}{4}}\left(\frac{t^2}{2}\right)+2 \Gamma \left(\frac{3}{4}\right) J_{\frac{3}{4}}\left(\frac{t^2}{2}\right)\right)}{\Gamma \left(\frac{1}{4}\right) J_{\frac{1}{4}}\left(\frac{t^2}{2}\right)-2 \Gamma \left(\frac{3}{4}\right) J_{-\frac{1}{4}}\left(\frac{t^2}{2}\right)},
\end{equation*}
where $\Gamma$ is the gamma function defined as,
\begin{equation*}
    \Gamma (z)=\int _{0}^{\infty} x^{z-1}e^{-x}\, \dd x
\end{equation*}
and $J$ is Bessel function of first kind defined as,
\begin{equation*}
    J_\nu(z) = \left(\frac{z}{2}\right)^\nu \sum_{k=0}^\infty \frac{\left(\frac{-z^2}{4}\right)^k}{k!\Gamma(\nu + k + 1)}
\end{equation*}

\begin{figure}[H]
	\centering\includegraphics[width=.5\linewidth]{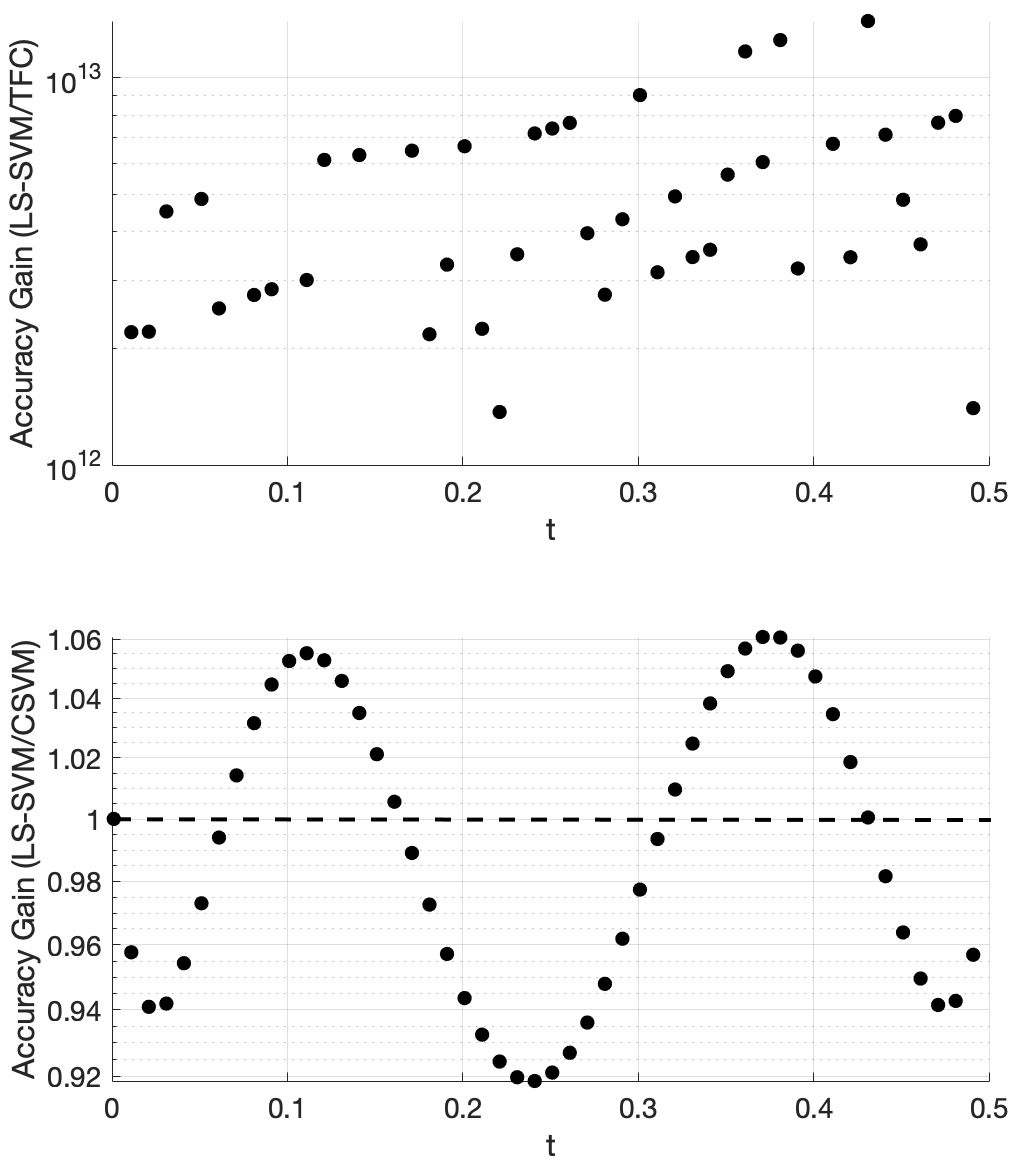}
    \caption{Accuracy gain for TFC and CSVM methods over LS-SVM for problem \#2 using 100 training~points.}
    \label{fig:Prob3}
\end{figure}

The accuracy gain of TFC and CSVM compared to LS-SVM for problem \#2 is shown in Figure~\ref{fig:Prob3}. This figure was created using 100 training points. The top plot shows the error in the LS-SVM solution divided by the TFC solution. The bottom plot provides the error in the LS-SVM solution divided by the error in the CSVM solution.
Figure \ref{fig:Prob3} shows that TFC is the most accurate of the two methods with the error being several orders of magnitude lower than the LS-SVM method. It was observed that the difference in accuracy between the CSVM and LS-SVM is negligible. The small variations in accuracy are a function of the specific method. For this problem, the solution accuracy for both methods monotonically decreases as $t$ increases; however, the behavior of this decrease is not constant and is at different rates, which produces a sine wave-like plot of the accuracy gain.

\begin{figure}[H]
	\centering\includegraphics[width=.5\linewidth]{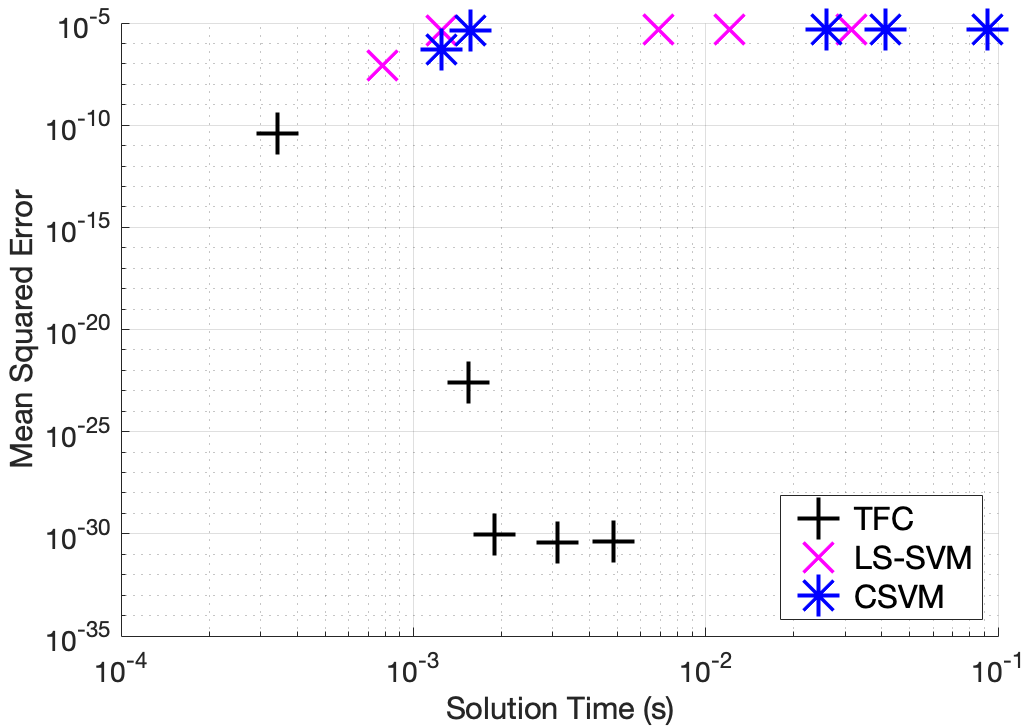}
    \caption{Mean squared error vs. solution time for problem \#2.}
    \label{fig:Prob3TimeAcc}
\end{figure}

Tables \ref{tab:TFC3} through \ref{tab:CSVM3} in the appendix compare the two methods for various numbers of training points when solving problem \#2. Additionally, these tables show that solving the DE using TFC is faster than using the LS-SVM method for all cases except the second case (using 16 training points). However, the speed gained using TFC is less than one order of magnitude. Furthermore, TFC is more accurate by multiple orders of magnitude as compared to the LS-SVM method over the entire range of test cases. In addition, TFC continues to reduce the MSE and maximum error on the test and training set as more training points are added, whereas the LS-SVM method error increases slightly between 8 and 16 points and then stays approximately the same. The CSVM method follows the same trend as the LS-SVM method; however, it requires more time to train than the LS-SVM method. This is highlighted in an accuracy versus speed comparison, shown graphically in Figure \ref{fig:Prob3TimeAcc}, where the MSE on the test set is plotted against training time for five specific cases: 8, 16, 32, 50, and 100 training points.

\subsection{Problem \#3} 
Problem \#3 is the second order linear ODE given by,
\begin{equation}\label{Eq:Prob4}
\begin{aligned}
    \ddot{y} + \frac{1}{5} \dot{y} + y =-\frac{1}{5} \, e^{-t/5} \cos t, \quad \text{subject to:} \begin{cases}y (0) = 1 \\ y' (0) = 1 \end{cases} \quad t \in [0,2],
\end{aligned}
\end{equation}
which has the analytic solution,
\begin{equation*}
    y(t) =  \frac{\sin(t)}{e^{t/5}}
\end{equation*}

\begin{figure}[H]
	\centering
    \includegraphics[width=.5\linewidth]{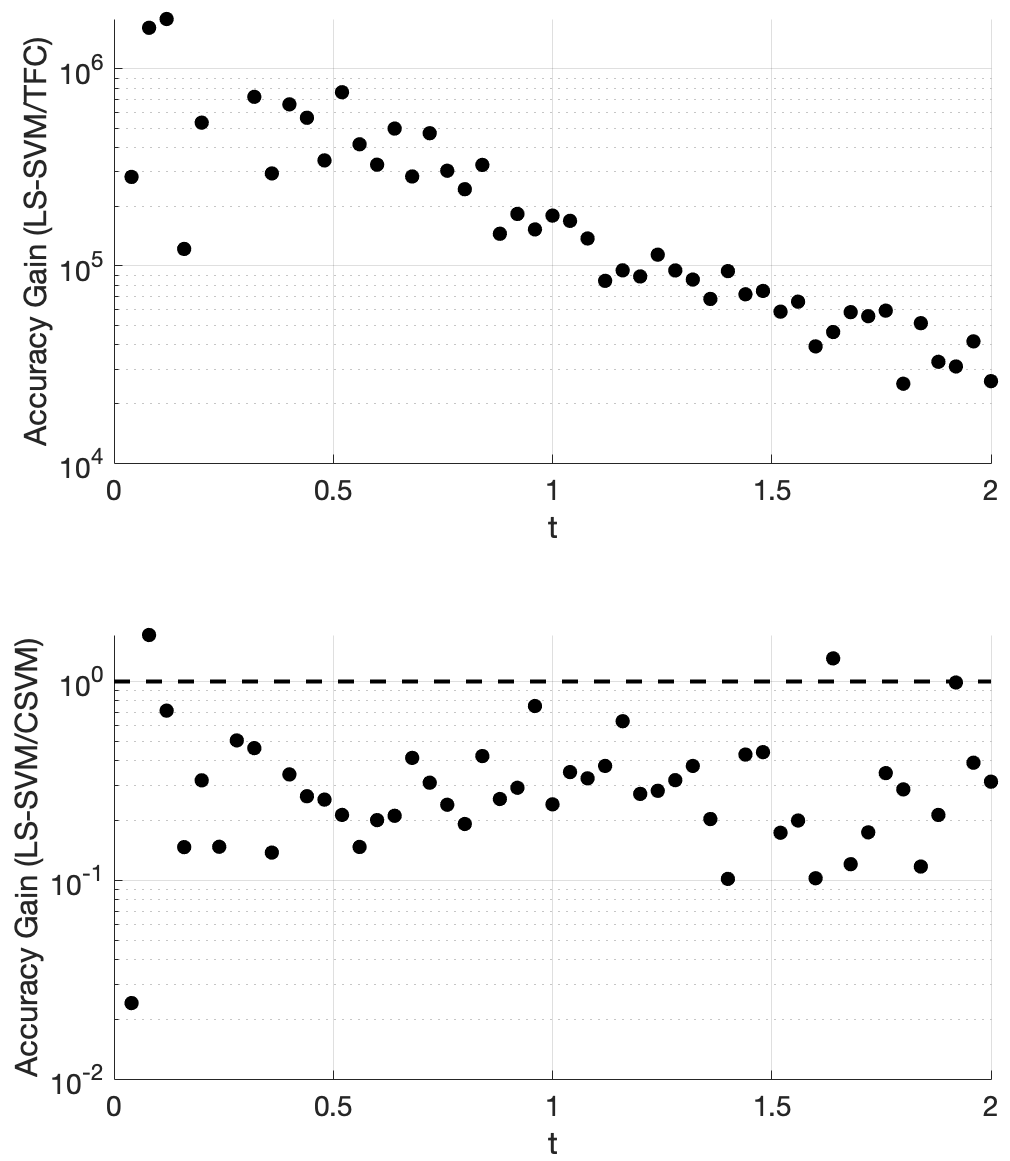}
    \caption{Accuracy gain for TFC and CSVM methods over LS-SVMs for problem \#3 using 100 training~points.}
    \label{fig:Prob4}
\end{figure}

The accuracy gain of TFC and CSVM compared to LS-SVM for problem \#3 is shown in Figure \ref{fig:Prob4}. The figure was created using 100 training points. The top plot shows the error in the LS-SVM solution divided by the TFC solution, and the bottom plot shows the error in the LS-SVM solution divided by the CSVM solution.

\begin{figure}[H]
	\centering\includegraphics[width=.5\linewidth]{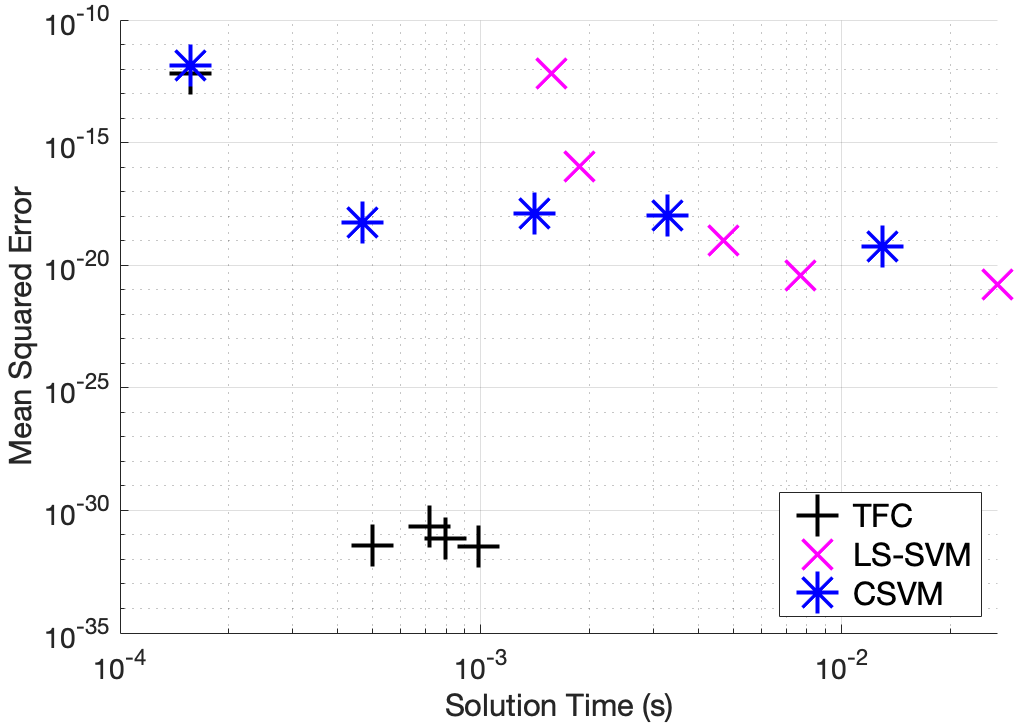}
    \caption{Mean squared error vs. solution time for problem \#3 accuracy vs. time.}
    \label{fig:Prob4TimeAcc}
\end{figure}

Tables \ref{tab:TFC4} through \ref{tab:CSVM4} in the appendix compare the two methods for various numbers of training points when solving problem \#3. These tables show that solving the DE using TFC is approximately an order of magnitude faster than using the LS-SVM method for all cases. Furthermore, TFC is more accurate than the LS-SVM method for all of the test cases. One interesting note is that when moving from 16 to 32 training points TFC actually loses a bit of accuracy, whereas the LS-SVM method continues to gain accuracy. Despite this, TFC is still multiple orders of magnitude more accurate than the LS-SVM method. Additionally, these tables show that the CSVM method is faster than the LS-SVM for all cases. The speed difference varies from approximately twice as fast to an order of magnitude faster. The LS-SVM and CSVM methods have a similar amount of error, and which method is more accurate depends on how many training points were being used. However, LS-SVM is slightly more accurate than CSVM for more cases than CSVM is slightly more accurate than LS-SVM. An accuracy versus speed comparison is shown graphically in Figure \ref{fig:Prob4TimeAcc}, where the MSE on the test set is plotted against training time for five specific cases: 8, 16, 32, 50, and 100 training points.

Figure \ref{fig:Prob4} shows that TFC is the most accurate of the three methods. The TFC error is 4 -- 6 orders of magnitude lower than the LS-SVM method. The LS-SVM method has error that is lower than the error in the CSVM method by an order of magnitude or less.
 
\subsection{Problem \#4} 

Problem \#4 is the second order linear PDE on $(x,y) \in [0,1]\times[0,1]$ given by,
\begin{equation}\label{Eq:Prob5}
   \nabla^2 z(x,y) = e^{-x}(x-2+y^3+6y) \quad \text{subject to:} \,
   \begin{cases}
   z(x,0) = xe^{-x}\\
   z(0,y) = y^3\\
   z(x,1) = e^{-x}(x+1)\\
   z(1,y) = (1+y^3)e^{-1}
   \end{cases}
\end{equation}
which has the analytical solution,
\begin{equation*}
    z(x,y) = (x + y^3) e^{-x}
\end{equation*}

\begin{figure}[H]
	\centering
    \includegraphics[width=.5\linewidth]{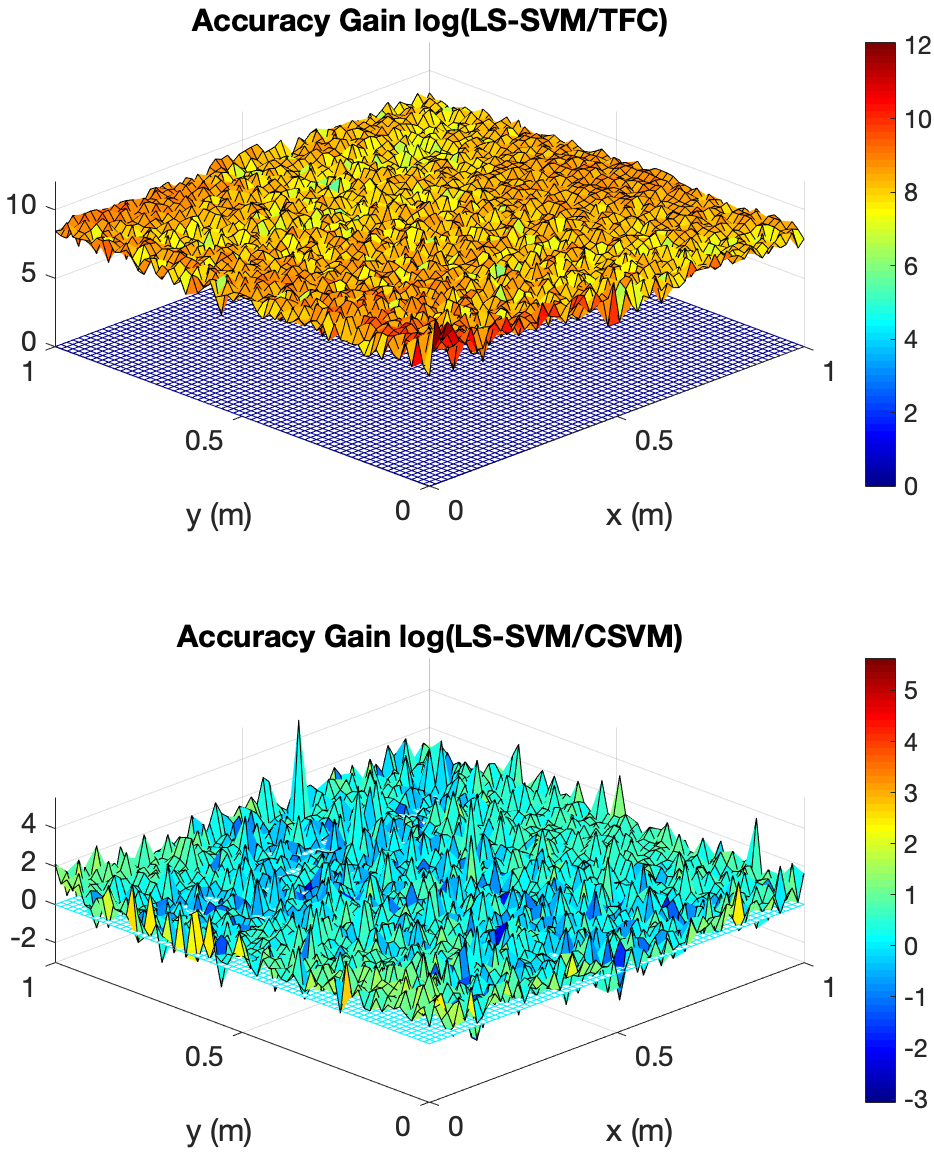}
    \caption{Accuracy gain for TFC and CSVM methods over LS-SVMs for problem \#4 using 100 training points in the domain.}
    \label{fig:Prob5}
\end{figure}

The accuracy gain of TFC and CSVM compared to LS-SVM for problem \#4 is shown in Figure \ref{fig:Prob5}. The figure was created using 100 training points in the domain---training points in the domain means training points that do not lie on one of the four boundaries. The top plot shows the log base 10 of the error in the LS-SVM solution divided by the TFC solution, and the bottom plot shows the log base 10 of the error in the LS-SVM solution divided by the error in the CSVM solution.
 
Figure \ref{fig:Prob5} shows that TFC is the most accurate of the two methods. The TFC error is orders of magnitude lower than the LS-SVM method. The CSVM error is, on average, approximately one order of magnitude lower than the LS-SVM method, but the error is still orders of magnitude higher than the error when using TFC. 
 
\begin{figure}[H]
	\centering\includegraphics[width=.5\linewidth]{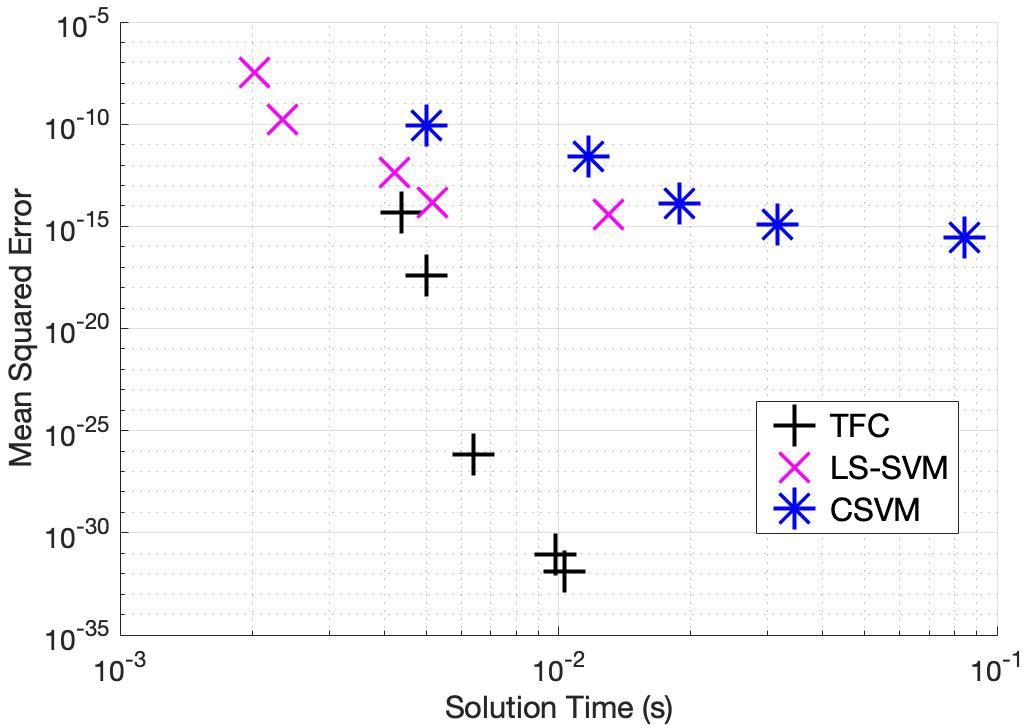}
    \caption{Mean squared error vs. solution time for problem \#4 accuracy vs. time.}
    \label{fig:Prob5TimeAcc}
\end{figure}

Tables \ref{tab:TFC5} through \ref{tab:CSVM5} in the appendix compare the two methods for various numbers of training points in the domain when solving problem \#4. These tables show that solving the DE using TFC is slower than LS-SVM by less than an order of magnitude for all test cases. The MSE error on the test set for TFC is less than LS-SVM for all of the test cases. The amount by which the MSE error on the test set differs between the two methods varies between 7 and 18 orders of magnitude. In addition, these tables show that the training time for CSVM is greater than LS-SVM by approximately an order of magnitude or less. The MSE error on the test set for CSVM is less than the MSE error on the test set for LS-SVM for all the test cases. The amount by which the MSE error on the test set differs between the two methods varies between one and three orders of magnitude. An accuracy versus speed comparison is shown graphically in Figure \ref{fig:Prob5TimeAcc}, where the MSE on the test set is plotted against training time for five specific cases: 9, 16, 36, 64, and 100 training points in the domain.

\section{Conclusion}
This article presented three methods to solve various types of DEs: TFC, LS-SVM, and CSVM. The CSVM method was a combination of the other two methods; it incorporated LS-SVM into the TFC framework. Four problems were presented that include a linear first order ODE, a nonlinear first order ODE, a linear second order ODE, and a linear second order PDE. The results showed that, in general, TFC is faster, by approximately an order of magnitude or less, and more accurate, by multiple orders of magnitude. The CSVM method has similar performance to the LS-SVM method, but the CSVM method satisfies the boundary constraints exactly whereas the LS-SVM method does not. While the CSVM method underperforms vanilla TFC, it showed the ease with which machine learning algorithms can be incorporated into the TFC framework. This capability is extremely important, as it provides a systematic way to analytically embed many different types of constraints into machine learning algorithms. 

This feature will be exploited in future studies and specifically for higher-dimensional PDEs, where the scalability of machine learning algorithms may give a major advantage over the orthogonal basis functions used in TFC. In this article, the authors found that the number of terms when using SVMs may become prohibitive at higher dimensions. Therefore, future work should focus on other machine learning algorithms, such as neural networks, that do not have this issue. Additionally, comparison problems should be looked at other than initial value problems. For example, problems could be used for comparison that have boundary value constraints, differential constraints, and integral constraints. 

Furthermore, future work should analyze the effect of the regularization term. One observation from the experiments is that the parameter $\gamma$ is very large, about $10^{13}$, making the contribution from $ \B{w}\T \B{w} $ insignificant. However, $ \B{w}\T \B{w} $ is the term meant to provide the best separating boundary surfaces. Going farther in this direction, one could analyze whether applying the kernel trick is beneficial (if the separating margin is not really achieved), or if an expansion similar to Chebyshev polynomials (CP), but using Gaussians, could provide a more accurate solution with a simpler algorithm. 

\vspace{6pt}

\funding{This work was supported by a NASA Space Technology Research Fellowship, Leake [NSTRF 2019] Grant \#: 80NSSC19K1152 and Johnston [NSTRF 2019] Grant \#: 80NSSC19K1149.}

\abbreviations{The following abbreviations are used in this manuscript:\\

\noindent 
\begin{tabular}{@{}ll}
BVP & boundary-value problem\\
CP & Chebyshev polynomial\\
CSVM & constrained support vector machines\\
DE & differential equation\\
IVP & initial-value problem\\
LS & least-squares\\
LS-SVM & least-squares support vector machines\\
MSE & mean square error\\
MVP & multi-value problem\\
ODE & ordinary differential equation\\
PDE & partial differential equation\\
RBF & radial basis function\\
SVM & support vector machines\\
TFC & \tfc
\end{tabular}}

\appendixtitles{no} 
\appendix
\section{Numerical Data}\label{apx:data}
The rows in Tables $\ref{tab:TFC1}$ through $\ref{tab:CSVM4}$ correspond to 8, 16, 32, 50, and 100 training points, respectively. 

\begin{table}[H]
\begin{center}
\caption{TFC results for problem \#1.}
\label{tab:TFC1}
\begin{tabular}{ccccccc}
\toprule
\makecell{\bf{Number of}\\\bf{Training}\\\bf{Points}} & \makecell{\bf{Training}\\\bf{Time (s)}} & \makecell{\bf{Maximum}\\\bf{Error on}\\\bf{Training Set}} & \makecell{\bf{MSE}\\\bf{on}\\\bf{Training Set}} & \makecell{\bf{Maximum}\\\bf{Error on}\\\bf{Test Set}} & \makecell{\bf{MSE}\\\bf{on}\\\bf{Test Set}} & \boldmath{$m$}\\
\midrule
8 & 7.813$\times 10^{-5}$ & 6.035$\times 10^{-6}$ & 1.057$\times 10^{-11}$ & 6.187$\times 10^{-6}$ & 8.651$\times 10^{-12}$ & 7 \\
16 & 1.406$\times 10^{-4}$ & 2.012$\times 10^{-11}$ & 1.257$\times 10^{-22}$ & 1.814$\times 10^{-11}$ & 8.964$\times 10^{-23}$ & 17 \\
32 & 5.000$\times 10^{-4}$ & 2.220$\times 10^{-16}$ & 1.887$\times 10^{-32}$ & 3.331$\times 10^{-16}$ & 2.086$\times 10^{-32}$ & 25 \\
50 & 7.500$\times 10^{-4}$ & 2.220$\times 10^{-16}$ & 9.368$\times 10^{-33}$ & 2.220$\times 10^{-16}$ & 1.801$\times 10^{-32}$ & 25 \\
100 & 1.266$\times 10^{-3}$ & 4.441$\times 10^{-16}$ & 1.750$\times 10^{-32}$ & 2.220$\times 10^{-16}$ & 1.138$\times 10^{-32}$ & 26 \\
\bottomrule
\end{tabular}
\end{center}
\end{table}

\begin{table}[H]
\begin{center}
\caption{{LS-SVM results for problem} \#1.}
\label{tab:SVM1}
\scalebox{0.95}{
\begin{tabular}{cccccccc}
\toprule
\makecell{\bf{Number of}\\\bf{Training}\\\bf{Points}} & \makecell{\bf{Training}\\\bf{Time (s)}} & \makecell{\bf{Maximum}\\\bf{Error on}\\\bf{Training Set}} & \makecell{\bf{MSE}\\\bf{on}\\\bf{Training Set}} & \makecell{\bf{Maximum}\\\bf{Error on}\\\bf{Test Set}} & \makecell{\bf{MSE}\\\bf{on}\\\bf{Test Set}} & \boldmath{$\gamma$} & \boldmath{$\sigma$}\\
\midrule
8 & 1.719$\times 10^{-3}$ & 1.179$\times 10^{-5}$ & 5.638$\times 10^{-11}$ & 1.439$\times 10^{-5}$ & 7.251$\times 10^{-11}$ & 5.995$\times 10^{17}$ & 3.162$\times 10^{0}$ \\
16 & 1.719$\times 10^{-3}$ & 1.710$\times 10^{-6}$ & 1.107$\times 10^{-12}$ & 1.849$\times 10^{-6}$ & 1.161$\times 10^{-12}$ & 3.594$\times 10^{15}$ & 6.813$\times 10^{-1}$ \\
32 & 2.188$\times 10^{-3}$ & 9.792$\times 10^{-8}$ & 3.439$\times 10^{-15}$ & 9.525$\times 10^{-8}$ & 3.359$\times 10^{-15}$ & 3.594$\times 10^{15}$ & 3.162$\times 10^{-1}$ \\
50 & 4.375$\times 10^{-3}$ & 1.440$\times 10^{-8}$ & 2.983$\times 10^{-17}$ & 8.586$\times 10^{-9}$ & 2.356$\times 10^{-17}$ & 3.594$\times 10^{15}$ & 3.162$\times 10^{-1}$ \\
100 & 1.031$\times 10^{-2}$ & 3.671$\times 10^{-9}$ & 3.781$\times 10^{-18}$ & 3.673$\times 10^{-9}$ & 3.947$\times 10^{-18}$ & 2.154$\times 10^{13}$ & 3.162$\times 10^{-1}$ \\
\bottomrule
\end{tabular}}
\end{center}
\end{table}

\begin{table}[H]
\begin{center}
\caption{CSVM results for problem \#1.}
\label{tab:SVMTFC1}
\scalebox{0.95}{
\begin{tabular}{cccccccc}
\toprule
\makecell{\bf{Number of}\\\bf{Training}\\\bf{Points}} & \makecell{\bf{Training}\\\bf{Time (s)}} & \makecell{\bf{Maximum}\\\bf{Error on}\\\bf{Training Set}} & \makecell{\bf{MSE}\\\bf{on}\\\bf{Training Set}} & \makecell{\bf{Maximum}\\\bf{Error on}\\\bf{Test Set}} & \makecell{\bf{MSE}\\\bf{on}\\\bf{Test Set}} & \boldmath{$\gamma$} & \boldmath{$\sigma$}\\
\midrule
8 & 3.125$\times 10^{-4}$ & 1.018$\times 10^{-5}$ & 4.131$\times 10^{-11}$ & 1.357$\times 10^{-5}$ & 5.547$\times 10^{-11}$ & 2.154$\times 10^{13}$ & 3.162$\times 10^{0}$ \\
16 & 1.406$\times 10^{-3}$ & 2.894$\times 10^{-7}$ & 2.588$\times 10^{-14}$ & 2.818$\times 10^{-7}$ & 2.468$\times 10^{-14}$ & 5.995$\times 10^{17}$ & 6.813$\times 10^{-1}$ \\
32 & 5.313$\times 10^{-3}$ & 2.283$\times 10^{-8}$ & 1.355$\times 10^{-16}$ & 2.576$\times 10^{-8}$ & 1.494$\times 10^{-16}$ & 3.594$\times 10^{15}$ & 3.162$\times 10^{-1}$ \\
50 & 3.281$\times 10^{-3}$ & 8.887$\times 10^{-9}$ & 2.055$\times 10^{-17}$ & 1.072$\times 10^{-8}$ & 2.783$\times 10^{-17}$ & 7.743$\times 10^{8}$ & 3.162$\times 10^{-1}$ \\
100 & 1.078$\times 10^{-2}$ & 2.230$\times 10^{-9}$ & 5.571$\times 10^{-19}$ & 2.163$\times 10^{-9}$ & 5.337$\times 10^{-19}$ & 3.594$\times 10^{15}$ & 1.468$\times 10^{-1}$ \\
\bottomrule
\end{tabular}}
\end{center}
\end{table}

\begin{table}[H]
\begin{center}
\caption{TFC results for problem \#2.}
\label{tab:TFC3}
\begin{tabular}{ccccccc}
\toprule
\makecell{\bf{Number of}\\\bf{Training}\\\bf{Points}} & \makecell{\bf{Training}\\\bf{Time (s)}} & \makecell{\bf{Maximum}\\\bf{Error on}\\\bf{Training Set}} & \makecell{\bf{MSE}\\\bf{on}\\\bf{Training Set}} & \makecell{\bf{Maximum}\\\bf{Error on}\\\bf{Test Set}} & \makecell{\bf{MSE}\\\bf{on}\\\bf{Test Set}} & \boldmath{$m$} \\
\midrule
8 & 3.437$\times 10^{-4}$ & 8.994$\times 10^{-6}$ & 2.242$\times 10^{-11}$ & 1.192$\times 10^{-5}$ & 4.132$\times 10^{-11}$ & 8 \\
16 & 1.547$\times 10^{-3}$ & 4.586$\times 10^{-12}$ & 6.514$\times 10^{-24}$ & 9.183$\times 10^{-12}$ & 2.431$\times 10^{-23}$ & 16 \\
32 & 1.891$\times 10^{-3}$ & 3.109$\times 10^{-15}$ & 9.291$\times 10^{-31}$ & 4.885$\times 10^{-15}$ & 9.590$\times 10^{-31}$ & 32 \\
50 & 3.125$\times 10^{-3}$ & 1.110$\times 10^{-15}$ & 2.100$\times 10^{-31}$ & 2.665$\times 10^{-15}$ & 3.954$\times 10^{-31}$ & 32 \\
100 & 4.828$\times 10^{-3}$ & 1.776$\times 10^{-15}$ & 3.722$\times 10^{-31}$ & 2.665$\times 10^{-15}$ & 4.321$\times 10^{-31}$ & 32 \\
\bottomrule
\end{tabular}
\end{center}
\end{table}

\begin{table}[H]
\begin{center}
\caption{LS-SVM results for problem \#2.}
\label{tab:SVM3}
\scalebox{0.95}{
\begin{tabular}{cccccccc}
\toprule
\makecell{\bf{Number of}\\\bf{Training}\\\bf{Points}} & \makecell{\bf{Training}\\\bf{Time (s)}} & \makecell{\bf{Maximum}\\\bf{Error on}\\\bf{Training Set}} & \makecell{\bf{MSE}\\\bf{on}\\\bf{Training Set}} & \makecell{\bf{Maximum}\\\bf{Error on}\\\bf{Test Set}} & \makecell{\bf{MSE}\\\bf{on}\\\bf{Test Set}} & \boldmath{$\gamma$} & \boldmath{$\sigma$} \\
\midrule
8 & 7.813$\times 10^{-4}$ & 1.001$\times 10^{-3}$ & 1.965$\times 10^{-7}$ & 1.001$\times 10^{-3}$ & 7.904$\times 10^{-8}$ & 1.000$\times 10^{10}$ & 3.704$\times 10^{-1}$ \\
16 & 1.250$\times 10^{-3}$ & 4.017$\times 10^{-3}$ & 4.909$\times 10^{-6}$ & 3.872$\times 10^{-3}$ & 4.514$\times 10^{-6}$ & 1.000$\times 10^{10}$ & 4.198$\times 10^{-1}$ \\
32 & 6.875$\times 10^{-3}$ & 4.046$\times 10^{-3}$ & 4.834$\times 10^{-6}$ & 3.900$\times 10^{-3}$ & 4.575$\times 10^{-6}$ & 1.000$\times 10^{10}$ & 4.536$\times 10^{-1}$ \\
50 & 1.203$\times 10^{-2}$ & 4.048$\times 10^{-3}$ & 4.792$\times 10^{-6}$ & 3.902$\times 10^{-3}$ & 4.580$\times 10^{-6}$ & 1.000$\times 10^{10}$ & 4.666$\times 10^{-1}$ \\
100 & 3.156$\times 10^{-2}$ & 4.050$\times 10^{-3}$ & 4.752$\times 10^{-6}$ & 3.903$\times 10^{-3}$ & 4.582$\times 10^{-6}$ & 1.000$\times 10^{10}$ & 4.853$\times 10^{-1}$ \\
\bottomrule
\end{tabular}}
\end{center}
\end{table}

\begin{table}[H]
\begin{center}
\caption{CSVM results for problem \#2.}
\label{tab:CSVM3}
\scalebox{0.95}{
\begin{tabular}{cccccccc}
\toprule
\makecell{\bf{Number of}\\\bf{Training}\\\bf{Points}} & \makecell{\bf{Training}\\\bf{Time (s)}} & \makecell{\bf{Maximum}\\\bf{Error on}\\\bf{Training Set}} & \makecell{\bf{MSE}\\\bf{on}\\\bf{Training Set}} & \makecell{\bf{Maximum}\\\bf{Error on}\\\bf{Test Set}} & \makecell{\bf{MSE}\\\bf{on}\\\bf{Test Set}} & \boldmath{$\gamma$} & \boldmath{$\sigma$} \\
\midrule
8 & 1.250$\times 10^{-3}$ & 1.556$\times 10^{-3}$ & 7.644$\times 10^{-7}$ & 1.480$\times 10^{-3}$ & 5.325$\times 10^{-7}$ & 1.000$\times 10^{10}$ & 3.452$\times 10^{-1}$ \\
16 & 1.563$\times 10^{-3}$ & 4.021$\times 10^{-3}$ & 4.914$\times 10^{-6}$ & 3.876$\times 10^{-3}$ & 4.517$\times 10^{-6}$ & 1.000$\times 10^{10}$ & 4.719$\times 10^{-1}$ \\
32 & 2.594$\times 10^{-2}$ & 4.047$\times 10^{-3}$ & 4.834$\times 10^{-6}$ & 3.901$\times 10^{-3}$ & 4.575$\times 10^{-6}$ & 1.000$\times 10^{10}$ & 5.109$\times 10^{-1}$ \\
50 & 4.109$\times 10^{-2}$ & 4.050$\times 10^{-3}$ & 4.792$\times 10^{-6}$ & 3.903$\times 10^{-3}$ & 4.580$\times 10^{-6}$ & 1.000$\times 10^{10}$ & 5.252$\times 10^{-1}$ \\
100 & 9.219$\times 10^{-2}$ & 4.051$\times 10^{-3}$ & 4.753$\times 10^{-6}$ & 3.904$\times 10^{-3}$ & 4.583$\times 10^{-6}$ & 1.000$\times 10^{10}$ & 5.469$\times 10^{-1}$\\
\bottomrule
\end{tabular}}
\end{center}
\end{table}

\begin{table}[H]
\begin{center}
\caption{TFC results for problem \#3.}
\label{tab:TFC4}
\begin{tabular}{cccccccc}
\toprule
\makecell{\bf{Number of}\\\bf{Training Points}} & \makecell{\bf{Training}\\\bf{Time (s)}} & \makecell{\bf{Maximum Error}\\\bf{on Training Set}} & \makecell{\bf{MSE on}\\\bf{Training Set}} & \makecell{\bf{Maximum Error}\\\bf{on Test Set}} & \makecell{\bf{MSE on}\\\bf{Test Set}} & \boldmath{$m$}\\
\midrule
8 & 1.563$\times 10^{-4}$ & 1.313$\times 10^{-6}$ & 5.184$\times 10^{-13}$ & 1.456$\times 10^{-6}$ & 6.818$\times 10^{-13}$ & 8 \\
16 & 7.969$\times 10^{-4}$ & 5.551$\times 10^{-16}$ & 6.123$\times 10^{-32}$ & 8.882$\times 10^{-16}$ & 7.229$\times 10^{-32}$ & 15 \\
32 & 7.187$\times 10^{-4}$ & 1.221$\times 10^{-15}$ & 2.377$\times 10^{-31}$ & 9.992$\times 10^{-16}$ & 2.229$\times 10^{-31}$ & 15 \\
50 & 5.000$\times 10^{-4}$ & 7.772$\times 10^{-16}$ & 3.991$\times 10^{-32}$ & 5.551$\times 10^{-16}$ & 3.672$\times 10^{-32}$ & 15 \\
100 & 9.844$\times 10^{-4}$ & 7.772$\times 10^{-16}$ & 5.525$\times 10^{-32}$ & 6.661$\times 10^{-16}$ & 3.518$\times 10^{-32}$ & 15 \\
\bottomrule
\end{tabular}
\end{center}
\end{table}

\begin{table}[H]
\begin{center}
\caption{LS-SVM results for problem \#3.}
\label{tab:SVM4}
\scalebox{0.9}{
\begin{tabular}{cccccccc}
\toprule
\makecell{\bf{Number of}\\\bf{Training Points}} & \makecell{\bf{Training}\\\bf{Time (s)}} & \makecell{\bf{Maximum Error}\\\bf{on Training Set}} & \makecell{\bf{MSE on}\\\bf{Training Set}} & \makecell{\bf{Maximum Error}\\\bf{on Test Set}} & \makecell{\bf{MSE on}\\\bf{Test Set}} & \boldmath{$\gamma$} & \boldmath{$\sigma$}\\
\midrule
8 & 1.563$\times 10^{-3}$ & 1.420$\times 10^{-6}$ & 8.300$\times 10^{-13}$ & 1.638$\times 10^{-6}$ & 6.522$\times 10^{-13}$ & 5.995$\times 10^{17}$ & 6.813$\times 10^{0}$ \\
16 & 1.875$\times 10^{-3}$ & 1.811$\times 10^{-8}$ & 1.015$\times 10^{-16}$ & 1.871$\times 10^{-8}$ & 1.014$\times 10^{-16}$ & 3.594$\times 10^{15}$ & 3.162$\times 10^{0}$ \\
32 & 4.687$\times 10^{-3}$ & 5.455$\times 10^{-10}$ & 1.025$\times 10^{-19}$ & 9.005$\times 10^{-10}$ & 1.015$\times 10^{-19}$ & 5.995$\times 10^{17}$ & 1.468$\times 10^{0}$ \\
50 & 7.656$\times 10^{-3}$ & 8.563$\times 10^{-11}$ & 3.771$\times 10^{-21}$ & 8.391$\times 10^{-11}$ & 3.646$\times 10^{-21}$ & 2.154$\times 10^{13}$ & 1.468$\times 10^{0}$ \\
100 & 2.688$\times 10^{-2}$ & 6.441$\times 10^{-11}$ & 1.500$\times 10^{-21}$ & 6.128$\times 10^{-11}$ & 1.640$\times 10^{-21}$ & 2.154$\times 10^{13}$ & 1.468$\times 10^{0}$ \\
\bottomrule
\end{tabular}}
\end{center}
\end{table}

\begin{table}[H]
\begin{center}
\caption{CSVM results for problem \#3.}
\label{tab:CSVM4}
\scalebox{0.9}{
\begin{tabular}{cccccccc}
\toprule
\makecell{\bf{Number of}\\\bf{Training Points}} & \makecell{\bf{Training}\\\bf{Time (s)}} & \makecell{\bf{Maximum Error}\\\bf{on Training Set}} & \makecell{\bf{MSE on}\\\bf{Training Set}} & \makecell{\bf{Maximum Error}\\\bf{on Test Set}} & \makecell{\bf{MSE on}\\\bf{Test Set}} & \boldmath{$\gamma$} & \boldmath{$\sigma$} \\
\midrule
8 & 1.563$\times 10^{-4}$ & 1.263$\times 10^{-6}$ & 7.737$\times 10^{-13}$ & 2.017$\times 10^{-6}$ & 1.339$\times 10^{-12}$ & 1.000$\times 10^{20}$ & 6.813$\times 10^{0}$ \\
16 & 4.687$\times 10^{-4}$ & 1.269$\times 10^{-9}$ & 4.961$\times 10^{-19}$ & 1.631$\times 10^{-9}$ & 5.342$\times 10^{-19}$ & 3.594$\times 10^{15}$ & 3.162$\times 10^{0}$ \\
32 & 1.406$\times 10^{-3}$ & 1.763$\times 10^{-9}$ & 8.308$\times 10^{-19}$ & 2.230$\times 10^{-9}$ & 1.248$\times 10^{-18}$ & 3.594$\times 10^{15}$ & 3.162$\times 10^{0}$ \\
50 & 3.281$\times 10^{-3}$ & 1.429$\times 10^{-9}$ & 1.045$\times 10^{-18}$ & 1.569$\times 10^{-9}$ & 1.017$\times 10^{-18}$ & 2.154$\times 10^{13}$ & 1.468$\times 10^{0}$ \\
100 & 1.297$\times 10^{-2}$ & 8.261$\times 10^{-10}$ & 8.832$\times 10^{-20}$ & 7.209$\times 10^{-10}$ & 5.589$\times 10^{-20}$ & 2.154$\times 10^{13}$ & 1.468$\times 10^{0}$ \\
\bottomrule
\end{tabular}}
\end{center}
\end{table}

The rows in tables $\ref{tab:TFC5}$ through $\ref{tab:CSVM5}$ correspond to 9, 16, 35, 64, and 100 training points, respectively. 

\begin{table}[H]
\begin{center}
\caption{TFC results for problem \#4.}
\label{tab:TFC5}
\begin{tabular}{cccccccc}
\toprule
\makecell{\bf{Number of}\\\bf{Training Points}\\\bf{in Domain}} & \makecell{\bf{Training}\\\bf{Time (s)}} & \makecell{\bf{Maximum Error}\\\bf{on Training Set}} & \makecell{\bf{MSE on}\\\bf{Training Set}} & \makecell{\bf{Maximum Error}\\\bf{on Test Set}} & \makecell{\bf{MSE on}\\\bf{Test Set}} & \boldmath{$m$}\\
\midrule
9 & 4.375$\times 10^{-3}$ & 1.107$\times 10^{-7}$ & 1.904$\times 10^{-15}$ & 1.543$\times 10^{-7}$ & 4.633$\times 10^{-15}$ & 8 \\
16 & 5.000$\times 10^{-3}$ & 3.336$\times 10^{-9}$ & 2.131$\times 10^{-18}$ & 4.938$\times 10^{-9}$ & 3.964$\times 10^{-18}$ & 9 \\
36 & 6.406$\times 10^{-3}$ & 6.628$\times 10^{-14}$ & 5.165$\times 10^{-28}$ & 2.333$\times 10^{-13}$ & 6.961$\times 10^{-27}$ & 12 \\
64 & 9.844$\times 10^{-3}$ & 4.441$\times 10^{-16}$ & 2.091$\times 10^{-32}$ & 8.882$\times 10^{-16}$ & 8.320$\times 10^{-32}$ & 15 \\
100 & 1.031$\times 10^{-2}$ & 3.331$\times 10^{-16}$ & 1.229$\times 10^{-32}$ & 6.661$\times 10^{-16}$ & 1.246$\times 10^{-32}$ & 15 \\
\bottomrule
\end{tabular}
\end{center}
\end{table}

\begin{table}[H]
\begin{center}
\caption{LS-SVM results for problem \#4.}
\label{tab:SVM5}
\scalebox{0.89}{
\begin{tabular}{cccccccc}
\toprule
\makecell{\bf{Number of}\\\bf{Training Points}\\\bf{in Domain}} & \makecell{\bf{Training}\\\bf{Time (s)}} & \makecell{\bf{Maximum Error}\\\bf{on Training Set}} & \makecell{\bf{MSE on}\\\bf{Training Set}} & \makecell{\bf{Maximum Error}\\\bf{on Test Set}} & \makecell{\bf{MSE on}\\\bf{Test Set}} & \boldmath{$\gamma$} & \boldmath{$\sigma$} \\
\midrule
9 & 2.031$\times 10^{-3}$ & 2.578$\times 10^{-4}$ & 9.984$\times 10^{-9}$ & 3.941$\times 10^{-4}$ & 3.533$\times 10^{-8}$ & 1.000$\times 10^{14}$ & 6.635$\times 10^{0}$ \\
16 & 2.344$\times 10^{-3}$ & 2.229$\times 10^{-5}$ & 6.277$\times 10^{-11}$ & 3.794$\times 10^{-5}$ & 1.731$\times 10^{-10}$ & 1.000$\times 10^{14}$ & 3.577$\times 10^{0}$ \\
36 & 4.219$\times 10^{-3}$ & 1.254$\times 10^{-6}$ & 2.542$\times 10^{-13}$ & 2.435$\times 10^{-6}$ & 4.517$\times 10^{-13}$ & 1.000$\times 10^{14}$ & 1.894$\times 10^{0}$ \\
64 & 5.156$\times 10^{-3}$ & 2.916$\times 10^{-7}$ & 1.193$\times 10^{-14}$ & 4.962$\times 10^{-7}$ & 1.390$\times 10^{-14}$ & 1.000$\times 10^{14}$ & 1.589$\times 10^{0}$ \\
100 & 1.297$\times 10^{-2}$ & 1.730$\times 10^{-7}$ & 3.028$\times 10^{-15}$ & 2.673$\times 10^{-7}$ & 3.668$\times 10^{-15}$ & 1.000$\times 10^{14}$ & 9.484$\times 10^{-1}$ \\
\bottomrule
\end{tabular}}
\end{center}
\end{table}

\begin{table}[H]
\begin{center}
\caption{CSVM results for problem \#4.}
\label{tab:CSVM5}
\scalebox{0.9}{
\begin{tabular}{cccccccc}
\toprule
\makecell{\bf{Number of}\\\bf{Training Points}\\\bf{in Domain}} & \makecell{\bf{Training}\\\bf{Time (s)}} & \makecell{\bf{Maximum Error}\\\bf{on Training Set}} & \makecell{\bf{MSE on}\\\bf{Training Set}} & \makecell{\bf{Maximum Error}\\\bf{on Test Set}} & \makecell{\bf{MSE on}\\\bf{Test Set}} & \boldmath{$\gamma$} & \boldmath{$\sigma$} \\
\midrule
9 & 5.000$\times 10^{-3}$ & 1.305$\times 10^{-5}$ & 1.936$\times 10^{-11}$ & 3.325$\times 10^{-5}$ & 8.262$\times 10^{-11}$ & 1.000$\times 10^{14}$ & 6.948$\times 10^{0}$ \\
16 & 1.172$\times 10^{-2}$ & 2.121$\times 10^{-6}$ & 7.965$\times 10^{-13}$ & 5.507$\times 10^{-6}$ & 2.530$\times 10^{-12}$ & 1.000$\times 10^{14}$ & 4.894$\times 10^{0}$ \\
36 & 1.891$\times 10^{-2}$ & 2.393$\times 10^{-7}$ & 6.242$\times 10^{-15}$ & 3.738$\times 10^{-7}$ & 1.341$\times 10^{-14}$ & 1.000$\times 10^{14}$ & 2.154$\times 10^{0}$ \\
64 & 3.156$\times 10^{-2}$ & 9.501$\times 10^{-8}$ & 1.021$\times 10^{-15}$ & 1.251$\times 10^{-7}$ & 1.165$\times 10^{-15}$ & 1.000$\times 10^{14}$ & 1.371$\times 10^{0}$ \\
100 & 8.453$\times 10^{-2}$ & 4.362$\times 10^{-8}$ & 2.687$\times 10^{-16}$ & 5.561$\times 10^{-8}$ & 2.951$\times 10^{-16}$ & 1.000$\times 10^{14}$ & 8.891$\times 10^{-1}$ \\
\bottomrule
\end{tabular}}
\end{center}
\end{table}

\section{Nonlinear ODE LS-SVM and CSVM Derivation}\label{app:B}
This appendix shows how the method of Lagrange multiplies is used to solve nonlinear ODEs using the LS-SVM and CSVM methods.
Equation (\ref{eq:LagNonLin}) shows the Lagrangian for the LS-SVM method.
\begin{figure*}[!h]
\begin{equation}\label{eq:LagNonLin}
\begin{aligned}
    {\cal L}(\B{w}, b, \B{e}, \B{y}, \B{\alpha}, \beta, \B{\eta}) = \frac{1}{2} (\B{w}\T \B{w} + \gamma \B{e}\T \B{e}) &-\sum_{i = 1}^N \alpha_i \left[\B{w}\T \B{\varphi}' (t_i) - f (t_i, y_i) - e_i\right] - \beta [\B{w}\T \B{\varphi} (t_0) + b - y_0] \\ &- \sum_{i=1}^N \eta_i \left[\B{w}\T \B{\varphi} (t_i) + b - y_i\right]
\end{aligned}
\end{equation}
\end{figure*}
The values where ${\cal L}$ are zero give candidates for the minimum.
\begin{align*}
    & \frac{\partial {\cal L}}{\partial \B{w}} = \B{0} \qquad\to\qquad \B{w} = \sum_{i=1}^N \alpha_i \B{\varphi}' (t_i) + \sum_{i=1}^N \eta_i \B{\varphi} (t_i) + \beta\B{\varphi} (t_0) \\
    &\frac{\partial {\cal L}}{\partial e_i}=0\qquad\to\qquad\gamma e_i=-\alpha_i\\
    &\frac{\partial {\cal L}}{\partial \alpha_i} = 0\qquad\to\qquad\B{w}\T\B{\varphi}'(t_i)=f(t_i,y_i)+e_i\\
    &\frac{\partial {\cal L}}{\partial \eta_i} = 0 \qquad\to\qquad y_i=\B{w}\T\B{\varphi}(t_i)+b\\
    &\frac{\partial {\cal L}}{\partial\beta} = 0 \qquad\to\qquad\B{w}\T\B{\varphi}(t_0)+b=y_0\\
    &\frac{\partial {\cal L}}{\partial b} = 0 \qquad\to\qquad\beta+\sum_{i=1}^N\eta_i=0\\
    &\frac{\partial {\cal L}}{\partial y_i} = 0 \qquad\to\qquad\alpha_i f_y(t_i,y_i)+\eta_i=0
\end{align*}

A system of equations can be set up by substituting the results found by differentiating ${\cal L}$ with respect to $\B{w}$ and $e_i$ into the remaining five equations found by taking the gradients of ${\cal L}$. This will lead to a set of $3N+2$ equations and $3N+2$ unknowns, which are $\alpha_i$, $\eta_i$, $y_i$, $\beta$, and $b$. This system of equations is given in Equation (\ref{eq:sys}),
\begin{equation}\label{eq:sys}
\begin{aligned}
    &\sum_{j=1}^N\alpha_j \B{\varphi}'(t_j)\T \B{\varphi}'(t_i) + \sum_{j=1}^N \eta_j \B{\varphi}(t_j)\T \B{\varphi}'(t_i) + \beta \B{\varphi}(t_0)\T \B{\varphi}'(t_i)+\frac{\alpha_i}{\gamma} = f(t_i,y_i) \\
    &\sum_{j=1}^N\alpha_j \B{\varphi}'(t_j)\T \B{\varphi}(t_i)+\sum_{j=1}^N \eta_j \B{\varphi}(t_j)\T \B{\varphi}(t_i) + \beta\B{\varphi}(t_0)\T\B{\varphi}(t_i)+b-y_i=0\\
    &\sum_{j=1}^N\alpha_j\B{\varphi}'(t_j)\T\B{\varphi}(t_0)+\sum_{j=1}^N \eta_j \B{\varphi}(t_j)\T \B{\varphi}(t_0) + \beta \B{\varphi}(t_0)\T\B{\varphi}(t_0)+b=y_0\\
    &\beta+\sum_{i=j}^N\eta_j=0\\
    &\alpha_if_y(t_i,y_i)+\eta_i=0\\
\end{aligned}
\end{equation}
where $i = 1,...,N$. The system of equations given in Equation (\ref{eq:sys}) is the same as the system of equations in Equation (20) of reference \cite{LS_SVP} with one exception: The regularization term, $I/\gamma$, in the second row of the second column entry is missing from this set of equations. The reason is, while running the experiments presented in this paper, that regularization term had an insignificant effect on the overall accuracy of the method. Moreover, as has been demonstrated here, it is not necessary in the setup of the problem. Once the set of equations has been solved, the model solution is given in the dual form~by,
\begin{equation*}
\begin{aligned}
    \hat{y}(t) =& \sum_{i=1}^N \alpha_i \B{\varphi}' (t_i)\T \B{\varphi} (t) + \sum_{i=1}^N \eta_i \B{\varphi} (t_i)\T \B{\varphi} (t) + \\
    & + \beta \B{\varphi} (t_0)\T \B{\varphi} (t) + b.
\end{aligned}
\end{equation*}

As with the linear ODE case, the set of $3N+2$ equations that need to be solved and dual form of the model solution can be written in terms of the kernel matrix and its derivatives. The method for solving the nonlinear ODEs with CSVM is the same, except the Lagrangian function is,
\begin{align*}
    {\cal L}(\B{w}, \B{e}, \B{y}, \B{\alpha}, \B{\eta}) = &\frac{1}{2} (\B{w}\T \B{w} + \gamma \B{e}\T \B{e}) -\sum_{i = 1}^N \alpha_i \left[\B{w}\T \B{\varphi}' (t_i) - f (t_i, y_i) - e_i\right] \\
    &- \sum_{i=1}^N \eta_i \left[\B{w}\T (\B{\varphi} (t_i)-\varphi(t_0)) + y_0 - y_i\right],
\end{align*}
where the initial value constraint has been embedded via TFC by taking the primal form of the solution to be,
\begin{equation*}
    \hat{y}(t) = \B{w}\T (\B{\varphi} (t_i)-\varphi(t_0)) + y_0.
\end{equation*}

Similar to the SVM derivation, taking the gradients of ${\cal L}$ and setting them equal to zero leads to a system of equations,
\begin{align*}
    &\sum_{j=1}^N \alpha_j\B{\varphi}^\prime(t_j)\T \B{\varphi}^\prime(t_i)+\sum_{j=1}^N \eta_j\Big(\B{\varphi}(t_j)-\B{\varphi}(t_0)\Big)\T \B{\varphi}\prime(t_i)-f(t_i,y_i)+\alpha_i/\gamma = 0\\
    &\sum_{j=1}^N \alpha_j\B{\varphi}^\prime(t_j)\T \Big(\B{\varphi}(t_i)-\B{\varphi}(t_0)\Big)+\sum_{j=1}^N \eta_j\Big(\B{\varphi}(t_j)-\B{\varphi}(t_0)\Big)\T\Big(\B{\varphi}(t_i)-\B{\varphi}(t_0)\Big)+y_0-y_i = 0\\
    &\alpha_if_y(t_i,y_i)+\eta_i = 0
\end{align*}
where $i = 1,...,N$. This can be solved using Newton's method for the unknowns $\alpha_i$, $\eta_i$, and $y_i$. In addition, the gradients can be used to re-write the estimated solution, $\hat{y}$, in the dual form,
\begin{equation*}
    \hat{y}(t) = \sum_{i=1}^N \alpha_i \B{\varphi}^\prime(t_i)\T\Big(\B{\varphi}(t)-\B{\varphi}(t_0)\Big)+\sum_{i=1}^N \eta_i\Big(\B{\varphi}(t_i)-\B{\varphi}(t_0)\Big)\T\Big(\B{\varphi}(t)-\B{\varphi}(t_0)\Big)+y_0.
\end{equation*}

As with the SVM derivation, the system of equations that must be solved and the dual form of the estimated solution can each be written in terms of the kernel matrix and its derivatives.

\section{Linear PDE CSVM Derivation}\label{app:C}
This appendix shows how to solve the PDE given by,
\begin{equation*}
   \nabla^2 z(x,y) = f(x,y) \quad \text{subject to:} \,
   \begin{cases}
   z(x,0) = c_1(x,0)\\
   z(0,y) = c_2(0,y)\\
   z(x,1) = c_3(x,1)\\
   z(1,y) = c_4(1,y)
   \end{cases}
\end{equation*}
using the CSVM method. Note, that this is the same PDE as shown in problem number four of the numerical results section where the right-hand side of the PDE has been replaced by a more general function $f(x,y)$ and the boundary-value type constraints have been replaced by more general functions $c_k(x,y)$ where $k=1,...,4$. Note, that throughout this section all matrices will be written using tensor notation rather than vector-matrix form for compactness. In this article, we will let superscripts denote a derivative with respect to the superscript variable and a subscript will be a normal tensor index. For example, the symbol $A^{xx}_{ij}$ would denote a second-order derivative of the second-order tensor $A_{ij}$ with respect to the variable $x$ (i.e. $\frac{\partial^2 A_{ij}}{\partial x^2}$).

Using the Multivariate TFC \cite{M-ToC}, the constrained expression for this problem can be written as,
\begin{align*}
    &\hat{z}(x,y) = A_{ij}v_iv_j + w_j\varphi_j(x,y) - w_kB_{ijk}v_iv_j\\
    &A_{ij} = \begin{bmatrix} 0 & c_1(x,0) & c_3(x,1) \\ c_2(0,y) & -c_1(0,0) & -c_3(0,1) \\ c_4(1,y) & -c_1(1,0) & -c_3(1,1)  \end{bmatrix} \\
    &B_{ijk} = \begin{bmatrix} 0 & \varphi_k(x,0) & \varphi_k(x,1) \\ \varphi_k(0,y) & -\varphi_k(0,0) & -\varphi_k(0,1) \\ \varphi_k(1,y) & -\varphi_k(1,0) & -\varphi_k(1,1) \end{bmatrix}\\
    &v_i = \begin{bmatrix} 1 & 1-x & x \end{bmatrix}\\
    &v_j = \begin{bmatrix} 1 & 1-y & y \end{bmatrix}.
\end{align*}
where $\hat{z}$ will satisfy the boundary constraints $c_k(x,y)$ regardless of the choice of $w$ and $\varphi_k$. Now, the Lagrange multiplies are added in to form ${\cal L}$,
\begin{equation*}
    {\cal L} (\B{w}, \B{\alpha}, \B{e}) = \frac{1}{2} w_i w_i + \frac{\gamma}{2} e_i e_i - \alpha_I (\hat{z}^{xx}_I+\hat{z}^{yy}_I-f_I-e_I),
\end{equation*}
where $\hat{z}_I$ is the vector composed of the elements $\hat{z}(x_n,y_n)$ where $n=1,...,N_p$ and there are $N_p$ training points.
The gradients of ${\cal L}$ give candidates for the minimum,
\begin{align*}
    &\frac{\partial \mathcal{L}}{\partial w_k} = w_k - \alpha_I(\varphi^{xx}_{Ik}-B^{xx}_{Iijk}v_iv_j+\varphi^{yy}_{Ik}-B^{yy}_{Iijk}v_iv_j) = 0\\
    & \frac{\partial{\cal L}}{\alpha_k} = \hat{z}^{xx}_I+\hat{z}^{yy}_I-f_I-e_I = 0 \\
    & \frac{\partial{\cal L}}{e_k} = \frac{\gamma}{2}e_k-\alpha_k = 0,
\end{align*}
where $\varphi_{Ik}$ is the second order tensor composed of the vectors $\varphi_i(x_n,y_n)$, $B_{Iijk}$ is the fourth order tensor composed of the third order tensors $B(x_n,y_n)_{ijk}$, and $n=1,...,N_p$. The gradients of ${\cal L}$ can be used to form a system of simultaneous linear equations to solve for the unknowns and write $\hat{z}$ in the dual form. The system of simultaneous linear equations is,
\begin{equation*}
    {\cal A}_{IJ}\alpha_J = {\cal B}_I \\
\end{equation*}
\begin{align*}
    {\cal A}_{IJ} = &\varphi^{xx}_{Ik}\varphi^{xx}_{Jk}-\varphi^{xx}_{Ik} B^{xx}_{Jijk}v_iv_j+\varphi^{xx}_{Ik}\varphi^{yy}_{Jk}-\varphi^{xx}_{Ik} B^{yy}_{Jijk}v_iv_j-B^{xx}_{Iijk}v_iv_j\varphi^{xx}_{Jk}+B^{xx}_{Iijk}v_iv_jB^{xx}_{Jmnk}v_mv_n\\
    &-B^{xx}_{Iijk}v_iv_j\varphi^{yy}_{Jk}+B^{xx}_{Iijk}v_iv_jB^{yy}_{Jmnk}v_mv_n+\varphi^{yy}_{Ik}\varphi^{xx}_{Jk}-\varphi^{yy}_{Ik} B^{xx}_{Jijk}v_iv_j+\varphi^{yy}_{Ik}\varphi^{yy}_{Jk}-\varphi^{yy}_{Ik} B^{yy}_{Jijk}v_iv_j\\
    &-B^{yy}_{Iijk}v_iv_j\varphi^{xx}_{Jk}+B^{yy}_{Iijk}v_iv_jB^{xx}_{Jmnk}v_mv_n-B^{yy}_{Iijk}v_iv_j\varphi^{yy}_{Jk}+B^{yy}_{Iijk}v_iv_jB^{yy}_{Jmnk}v_mv_n+\frac{1}{\gamma}\delta_{IJ}
\end{align*}
\begin{equation*}
    {\cal B}_I = f_I-A^{xx}_{Iij}v_iv_j-A^{yy}_{Iij}v_iv_j
\end{equation*}
where $v_m=v_i$, $v_n=v_j$, and $A_{Iijk}$ is the fourth order tensor composed of the third order tensors $A(x_n,y_n)_{ijk}$ where $n=1,...,N_p$. The dual-form of the solution is,
\begin{align*}
    \hat{z}(x,y) = &A_{ij}v_iv_j +\alpha_I\bigg[\varphi^{xx}_{Ik}\varphi(x,y)_k-B^{xx}_{Iijk}v_iv_j\varphi_k(x,y)+\varphi^{yy}_{Ik}\varphi_k(x,y)-B^{yy}_{Iijk}v_iv_j\varphi_k(x,y)\bigg]\\
    &-\alpha_I\bigg[\varphi^{xx}_{Ik}B_{ijk}v_iv_j-B^{xx}_{Iijk}v_iv_jB_{mnk}v_mv_n+\varphi^{yy}_{Ik}B_{ijk}v_iv_j-B^{yy}_{Iijk}v_iv_jB_{mnk}v_mv_n\bigg].
\end{align*}

The system of simulatenous linear equations as well as the dual form of the solution can be written and were solved using the kernel matrix and its partial derivatives.

\bibliographystyle{unsrtnat}
\bibliography{mybib}


\end{document}